# A Comparative Investigation of Compositional Syntax and Semantics in DALL·E 2

Elliot Murphy[1,2*], Jill de Villiers[3], Sofia Lucero Morales[3]

1. Vivian L. Smith Department of Neurosurgery, University of Texas Health Science Center at Houston, TX, USA
2. Texas Institute for Restorative Neurotechnologies, University of Texas Health Science Center at Houston, TX, USA
3. Department of Psychology, Smith College, MA, USA

*Correspondence should be addressed to elliot.murphy@uth.tmc.edu

**Abstract**: In this study we compared how well DALL·E 2 visually represented the meaning of linguistic prompts also given to young children in comprehension tests. Sentences representing fundamental components of grammatical knowledge were selected from assessment tests used with several hundred English-speaking children aged 2–7 years for whom we had collected original item-level data. DALL·E 2 was given these prompts five times to generate 20 cartoons per item, for 9 adult judges to score. Results revealed no conditions in which DALL·E 2-generated images that matched the semantic accuracy of children, even at the youngest age (2 years). DALL·E 2 failed to assign the appropriate roles in reversible forms; it failed on negation despite an easier contrastive prompt than the children received; it often assigned the adjective to the wrong noun; it ignored implicit agents in passives. This work points to a clear absence of compositional sentence representations for DALL·E 2.

**Introduction**
DALL·E 2 and 3 are foundation text-to-image models that have the ability to turn text prompts into images (Ramesh et al. 2022). Created by OpenAI, DALL·E 2/3 generates novel synthetic images corresponding to input text, often with remarkable exactness. Through its deep learning architecture, this program has been previously described as comprehending the natural language prompts it receives and relating their components "by understanding the verbs that connect them" (Kargwal 2022). However, previous work indicates limits to DALL·E 2's ability to handle syntactic information (Leivada et al. 2023), suggesting that any success on syntax tasks may be partly illusory and due to semantic keyword searches (Dentella et al. 2023b, Marcus et al. 2022, 2023). The most recent research has explored the (in)ability of DALL·E 2 and Midjourney to count objects (Wasielewski 2023), but human language utilizes processes that go beyond counting and linearity. Compositional syntax-semantics is often thought to be the defining feature of the human language faculty (Marcolli et al. 2023, Murphy 2024a, 2024b, Murphy et al. In press). The considerable performance variation across languages other than English for DALL·E 2 (Reviriego & Merino-Gómez 2022) already points towards the potential lack of a fundamental compositional strategy to represent structured linguistic meaning. To further test DALL·E 2's abilities, we compared how well children in comprehension tests understood the meaning of the same linguistic prompts we gave to DALL·E 2, to develop a comparable benchmark. To our knowledge this constitutes the first direct comparative test of DALL·E 2's abilities and children's abilities using the *same linguistic items*. Sentences were selected from assessment tests (Golinkoff et al. 2017, Jackson et al. 2023, Seymour et al. 2005) used with several hundred English-speaking children aged 2–7 years for whom we had original item-level data.

In this study, we focused on foundational aspects of grammar:

1) A basic operation required for **propositional semantics** is establishing the roles of agent and patient in a declarative sentence. This is tested by using reversible transitives for which the objects named by the noun could play either role with a chosen verb ('The car hit the truck').
2) **Negation** is central to syntax-semantics ('One girl has shoes, one girl has no shoes').
3) **Prepositions** can link nouns together, where the relation and order matter ('The book is on the cup').
4) Embedded adjectives need to have appropriately **assigned scope**, e.g. 'The kitten in a cup with a yellow ribbon' does not mean the cup or the kitten is yellow.
5) The **passive voice** entails more than reversed roles ('The cat is being dressed' implies the existence of an external agent) (Roeper 2004).

**Methods**
We prompted DALL·E 2 with a sentence 5 times to generate 20 images each for an online survey for 9 adult judges (university educated). 'Cartoon style' was specified to limit the possibility that DALL·E 2 could find online photographs with captions matching prompts. The DALL·E 2 searches were run in May 2023. Several searches of similar commands were run to replicate the results (see Appendix II), crucially with no filtering or selection procedure. This method helps to extend the approach reported in Leivada et al. (2023), which did not solicit external, non-expert (hence, non-biased) judges for assessing image outputs.

The 9 volunteer judges from different disciplines completed Google forms (560 images) to judge on a 3-point scale whether the images were (1) accurate depictions, (2) partially accurate but containing a misleading feature, or were (3) inaccurate (Figure 1). The average score provided an adequacy percentage, such that, for example, an average of 3 would yield 0% adequacy, an average of 2 would yield 50% adequacy, and an average of 1.5 would yield 75% adequacy. A second index considered only the scores of 3, and counted percentage correct. The percentage correct is the index most naturally compared to the children's scores, but we considered both.

Exact instructions provided to our judges are given in Appendix I. We also show extensive examples from negation, prepositional structures and transitive structures (Appendix II). Querying text-to-image applications in order to examine which properties of the text can be successfully captured and represented is an error-proof method that needs little defense, in particular given that we test the application in the domain that it has been tailored to perform best: transforming natural language prompts into realistic images.

Data from children across various ages (2;0–6;11) were taken from sources providing large behavioral samples for performance on each prompt. These were mostly picture choice tasks from large language assessments samples (QUILS-TOD, Jackson et al. 2023; QUILS, Golinkoff et al. 2017; DELV, Seymour et al. 2005), with the exception of one act-out task (Chan et al. 2010). Percentage correct was taken as the comparable score for children. Importantly, while children had to select from multiple pictured options, DALL·E 2 had to create a novel image, generating a divergence in typical task demands and processing strategies, yet we argue that the underlying demand for a compositional meaning was salient in both tasks. Given that is it not feasible to ask 2-3-year-old children to draw referents for even simple linguistic structures, this was an unavoidable difference. Due to the potentially misleading nature of direct statistical comparisons between divergent tasks, we report here grouped averages (Figures 5-9), *à la* Leivada et al. (2023).

**Results**
Results revealed no conditions in which DALL·E 2-generated images that matched the semantic accuracy of children, even at the youngest age (2 years) (Figure 4). DALL·E 2 failed to assign the appropriate roles in reversible transitive and prepositional phrase forms; it fared poorly on negation despite an easier contrastive prompt[1] than the children received (Figure 2); it often assigned the adjective to the wrong noun; it ignored implicit agents in passives (Figure 2, 3). Across the board, even children as young as 2 years-old outperformed DALL·E 2 (Figures 5-10).

---

[1] Children were asked to choose from a set of boys: "Find the boy with no shoes". We asked DALL·E 2 to depict: "One boy has shoes, one boy has no shoes". This was necessary to test whether the negation was being attended to at all.

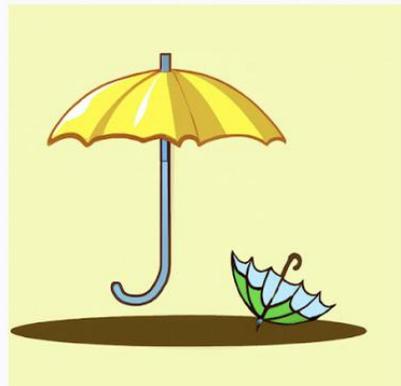

**"The umbrella is below the swing"**

Do you think this image is
(1) accurate
(2) includes a potentially misleading feature
(3) inaccurate

Explain your reason below:

____________________________________

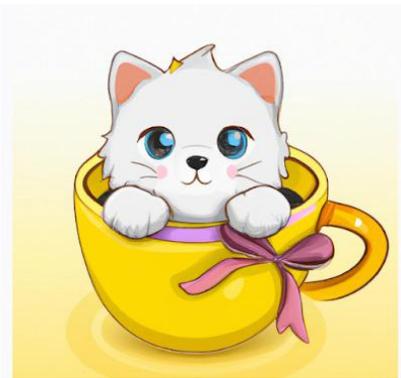

**"The kitten is in a cup with a yellow ribbon"**

Do you think this image is
(1) accurate
(2) includes a potentially misleading feature
(3) inaccurate

Explain your reason below:

____________________________________

**Figure 1**: Example questions for the adult judges based on example DALL·E 2 generated images shown along with the text prompt used to generate the image.

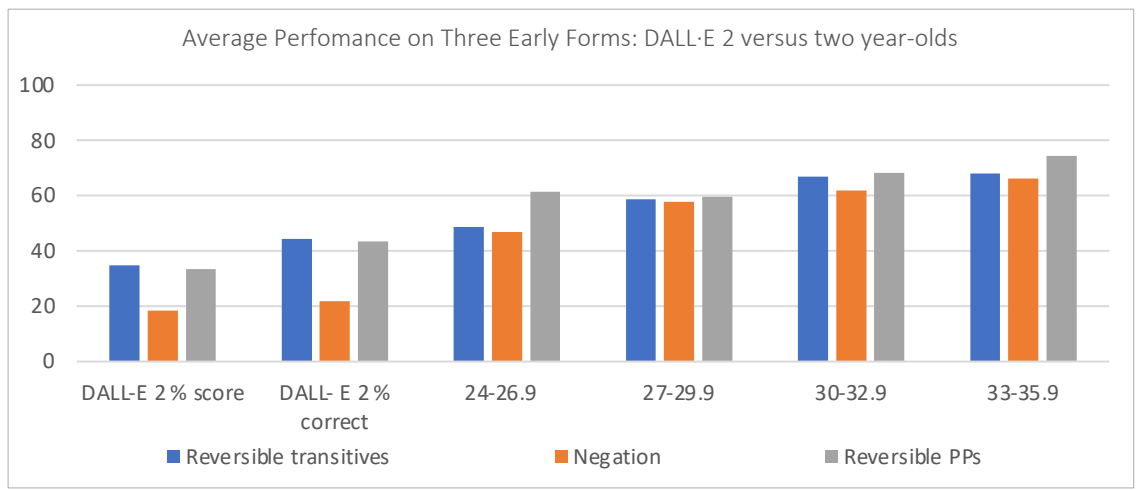

**Figure 2**: The performance of DALL·E 2 compared to children aged 24-35 months-old (age is in months). Percentage score = adequacy metric; percentage correct = accuracy metric (see Methods). Y-axis = percentage correct (for examples, see Appendix II).

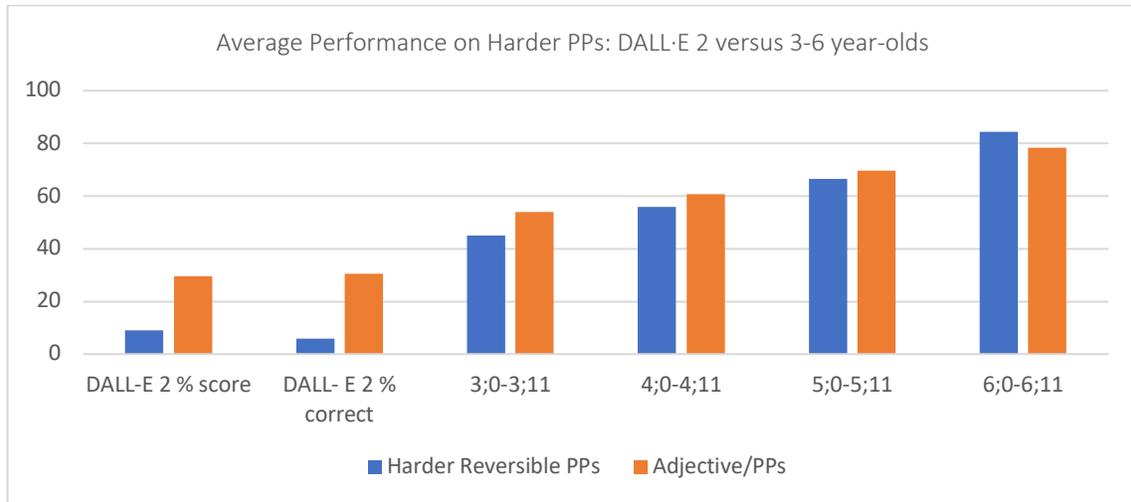

**Figure 3**: The performance of DALL·E 2 compared to children aged 3-6 years (age range is in youngest years;months – oldest years;months). Y-axis = percentage correct.

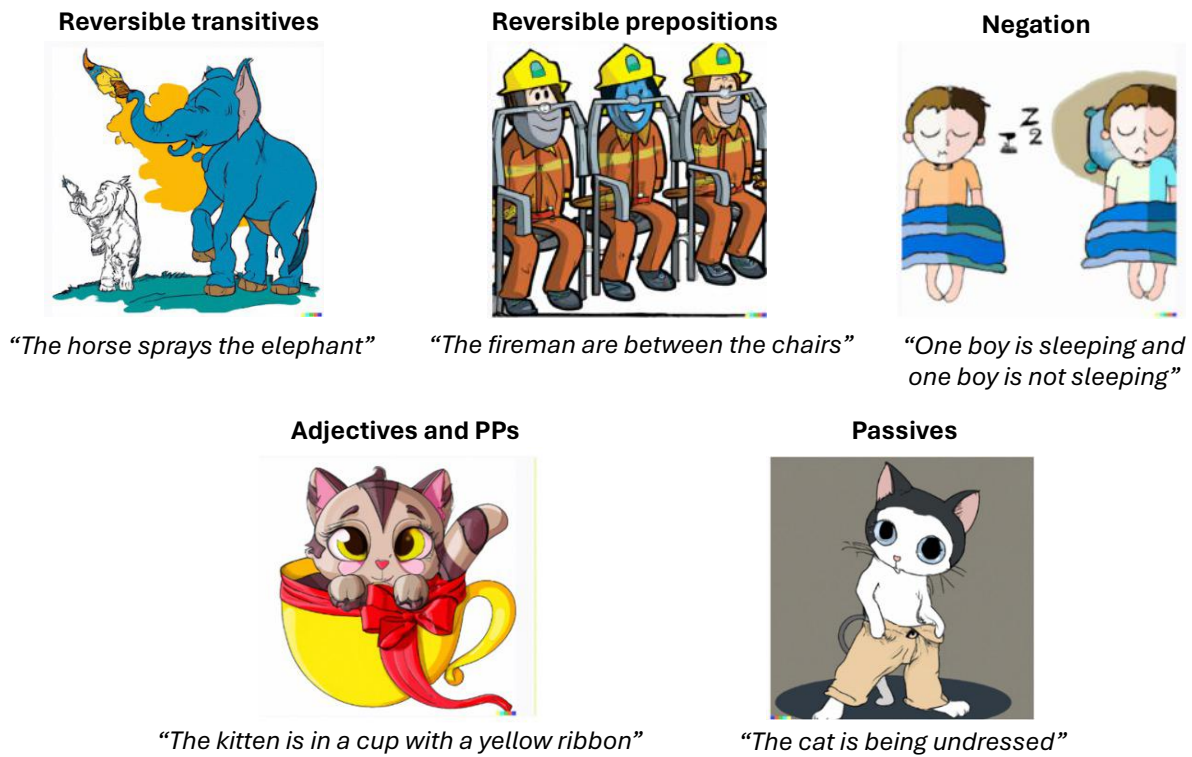

**Figure 4**: The performance of DALL·E 2 on a range of compositional syntactic processes (see Appendix II). Precise prompts are provided below example images generated by DALL·E 2.

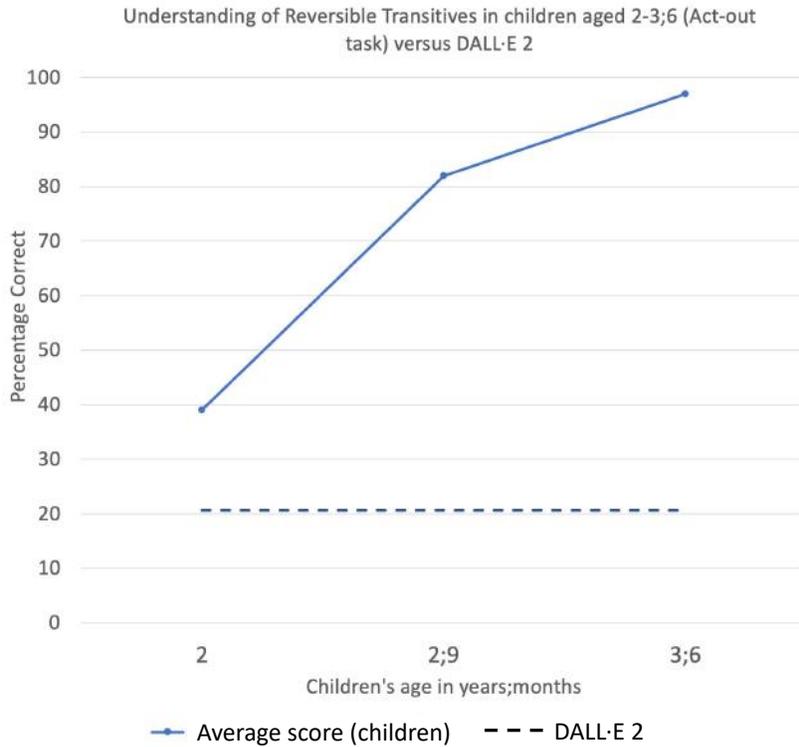

**Figure 5**: The performance of DALL·E 2 vs. children tasked with acting out the sentence meaning of reversible transitive sentences. We note that Chan et al. (2010) provided items but not item-level data, so data has been averaged for comparison.

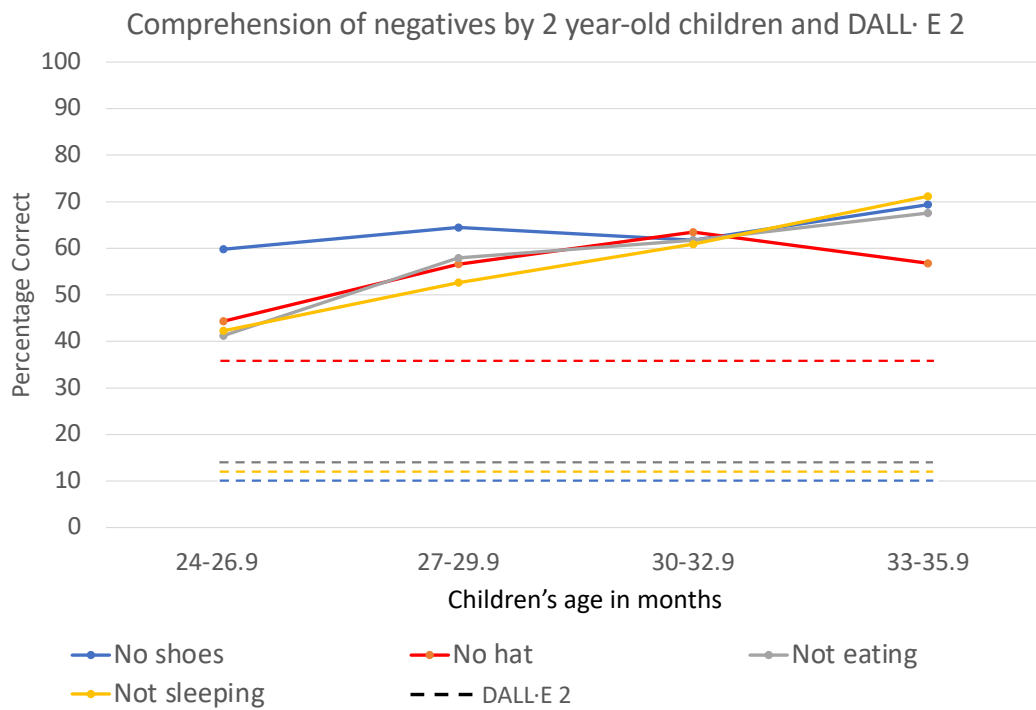

**Figure 6**: The performance of DALL·E 2 vs. children on negatives. Dotted lines represent DALL·E 2 performance on correspondingly colored items.

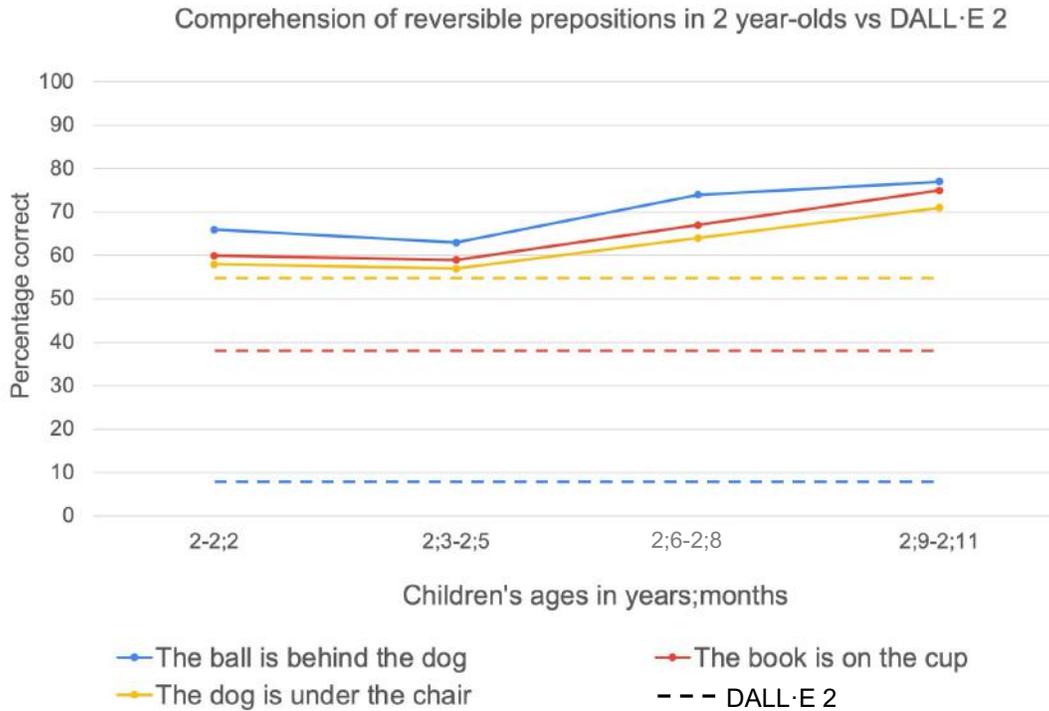

**Figure 7**: The performance of DALL·E 2 vs. children on reversible prepositions. Dotted lines represent DALL·E 2 performance on correspondingly colored items.

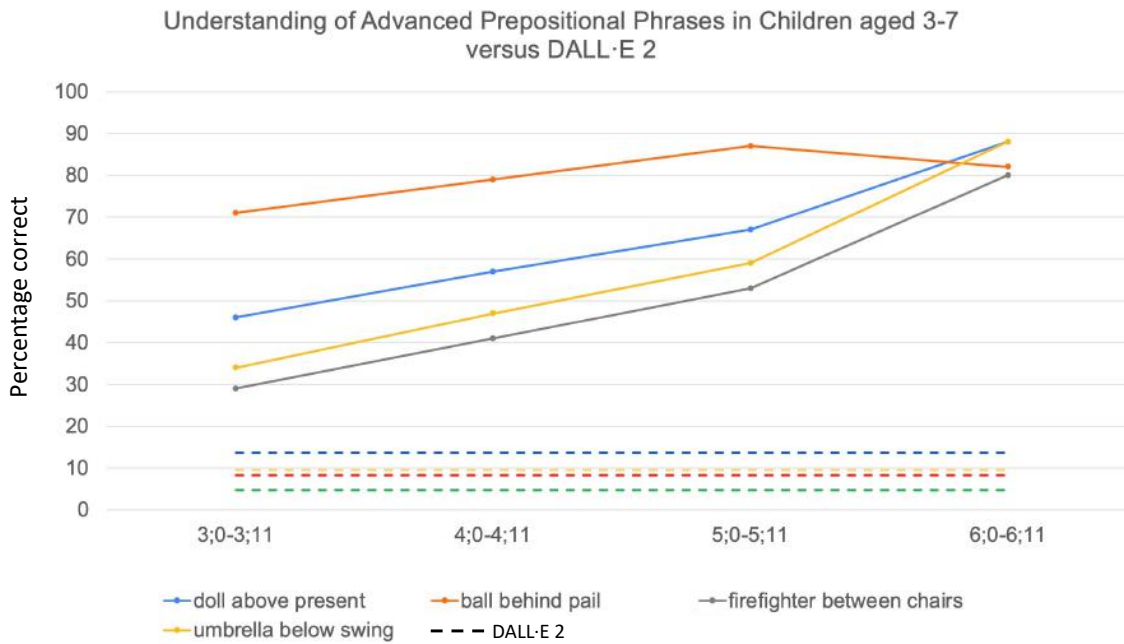

**Figure 8**: The performance of DALL·E 2 vs. children on complex prepositional phrases. Dotted lines represent DALL·E 2 performance on correspondingly colored items.

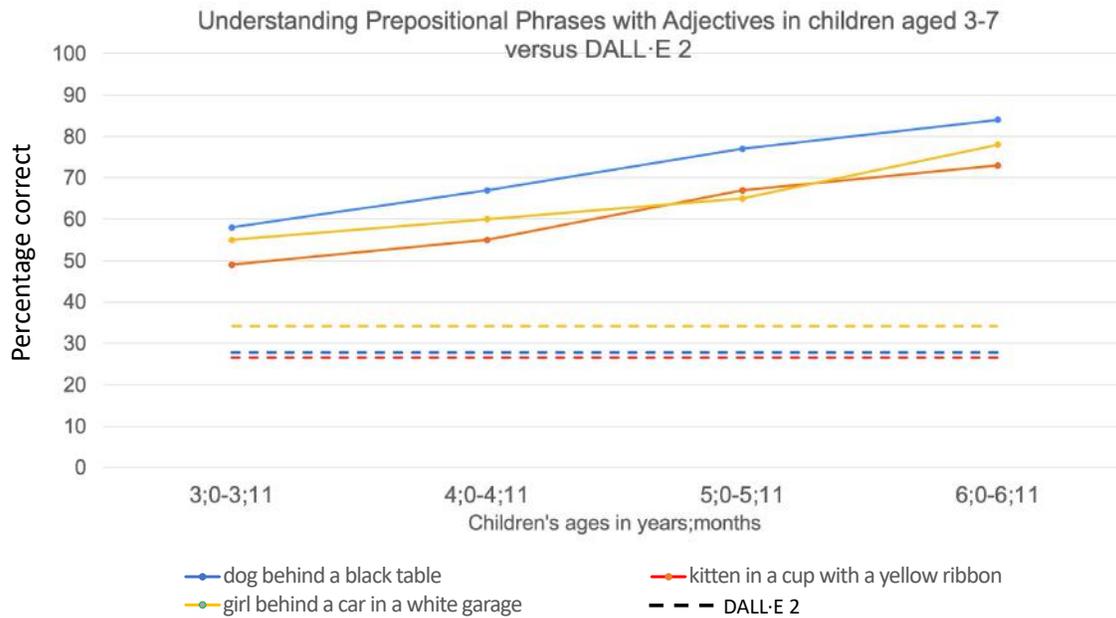

**Figure 9**: The performance of DALL·E 2 vs. children on prepositional phrases with adjectives. Dotted lines represent DALL·E 2 performance on correspondingly colored items.

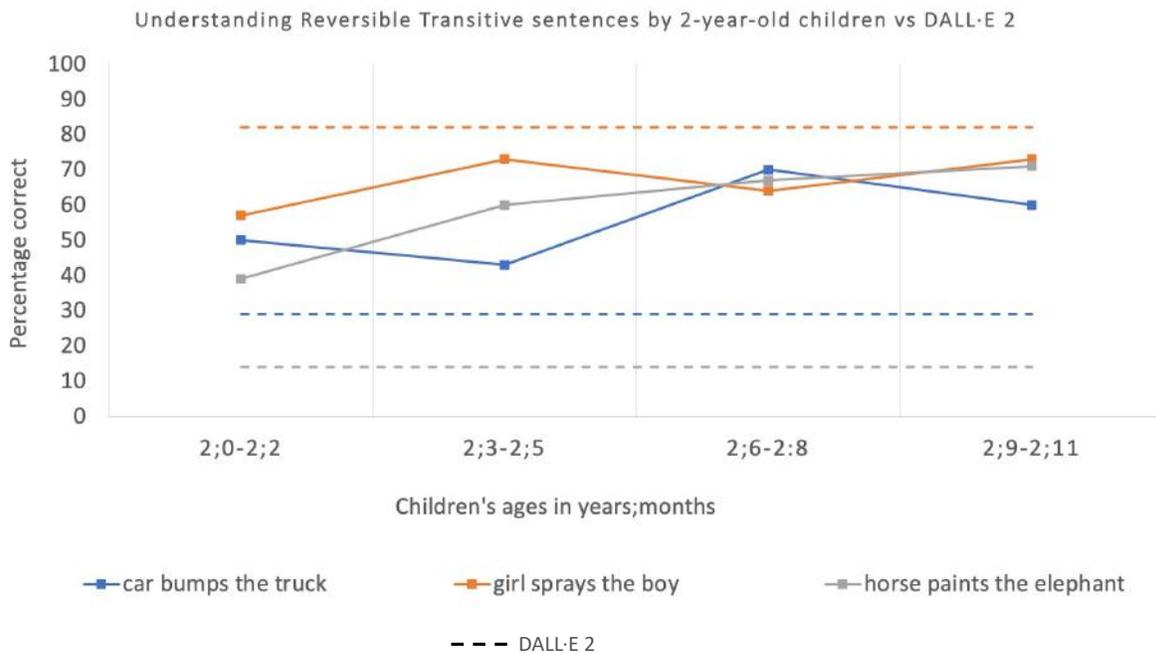

**Figure 10**: The performance of DALL·E 2 vs. children on reversible transitives. Dotted lines represent DALL·E 2 performance on correspondingly colored items.

# Discussion

This study provides the first direct comparison between child language comprehension and artificial text-to-image models. Our work points towards an absence of compositional sentence representations for DALL·E 2. In some of the cases we presented, children's expertise in core knowledge of objects seems to contribute to their higher performance scores. For example, consider the example (Figure 4) of "the firemen are between the chairs". DALL·E 2 in this and other cases did not respect ordinary assumptions about human bodies in space, but instead wove the firemen and chairs together. This kind of error goes beyond adding extra digits to hands, or the statistical bias of representing women rather than men holding babies, and represents a fundamental error in compositional syntax-semantics. Human children are able to construct a grammar with direct links to compositional meaning, connecting language to internal cognitive models of the world – an architecture that seems absent in DALL·E 2, which does not seem able to reliably and robustly construct linguistically coherent representations.

We confirm and extend the findings from Leivada et al. (2023) that a fundamental aspect of human language – compositionality – is lacking from DALL·E 2 (see Murphy 2023, Murphy & Leivada 2022). This feature is present early and robustly in human children (Perkins & Lidz 2021). Our work also dovetails into recent concerns about the lack of grammatical competence for language models, and the presence of certain response biases that deviate from standard assumptions concerning common processing strategies that humans possess for language (Dentella et al. 2023a). The importance of syntax-specific inductive biases cannot be underestimated (Sartran et al. 2022), in particular given that grammar reflects the organization of a specific mode of thought that appears to be unique to humans (Hinzen 2006, McCarty et al. 2023, Murphy 2015, 2020, 2024a, Murphy et al., 2022). We also note the promising direction that neurosymbolic approaches are offering the field (Trinh et al. 2024), which seem ripe for adaptation towards syntactic representations given their moves towards cognitively plausible architectures constituting a hybrid of rich prior knowledge and sophisticated reasoning techniques (Marcus 2020). In addition to grammar, the pragmatics (and 'super pragmatics'; Schlöder & Altshuler 2023) of language processing also appears absent in current artificial models. Future research should explore additional aspects of functional grammatical structure to test these limits further (Sonkar et al. 2022), given the centrality of function words in regulating syntactic information (for broader discussion, see Merrill et al. 2024).

Future work could seek to approximate the type of task conducted by both DALL·E 2/3 and child participants, to make the task demands as similar as possible, given that in our comparison, children had to select from multiple options whereas DALL·E 2 had to create a novel image. Nevertheless, despite divergent task demands, the stark accuracy differences reported here across all of our linguistic structures provides strong evidence for the absence of a higher-order compositional apparatus for DALL·E 2, which forms the fundamental basis of human language (Partee 1984, Pietroski 2018).

While previous work has made clear that DALL·E 2 does not rise to the level of adult compositional competence, we have shown here that it also does not rise even to the level of a 2-3-year-old's linguistic competence.

**Conflict of interest**: The authors declare that they have no conflict of interest.

**Appendix I**

Instructions provided to adult judges:

*We are testing whether DALL·E can respond to a prompt by drawing a cartoon picture that matches it closely in essential details. We are interested in whether the essential components of the meaning of the prompt are captured, or whether DALL-E fails to capture all components. We are not so much interested in small glitches (e.g., a three armed woman) or physical details of the drawings (in reality, that tree would fall over) as in whether the essence of the linguistic meaning is captured. Are the right characters involved, and are they doing what the verb specifies? However, sometimes it is hard to tell these apart. For example, it might be that case that DALL-E provides an image that could possibly be seen as capturing the meaning of the prompt, but that some other important details might be missing.*

*A simple example might be the prompt "A man looking at a lion through a telescope", but the image provided only shows a man looking through a telescope, with no lion on the screen. Technically, the man could be looking at a lion, but he could also be looking at something else entirely, and so this example would not fully capture the right meaning.*

*We provide twenty attempts at each prompt. Please indicate below each as follows:*

*(1) for the photo that matches the essentials of the prompt's linguistic meaning*

*(2) for a photo that matches the prompt's linguistic meaning but has some peculiarities in the depiction (please specify)*

*(3) for an image that is an impossible match for the prompt, e.g., reverses the meaning.*

**Appendix II**

1. **Negation examples**

***One girl has shoes and one girl has no shoes***

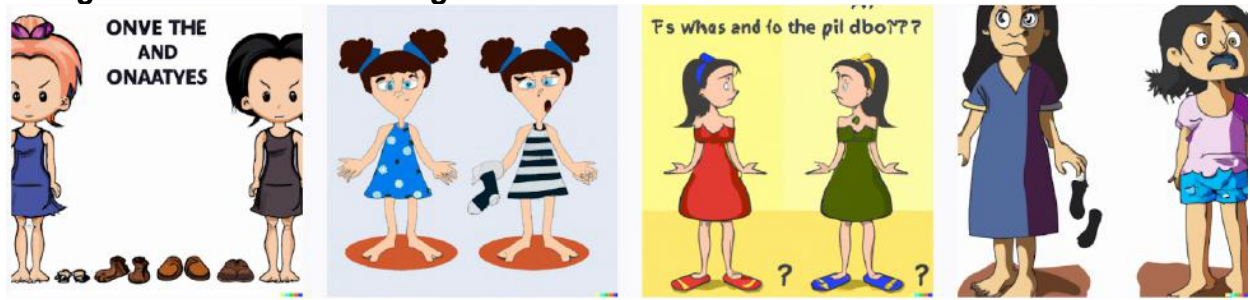

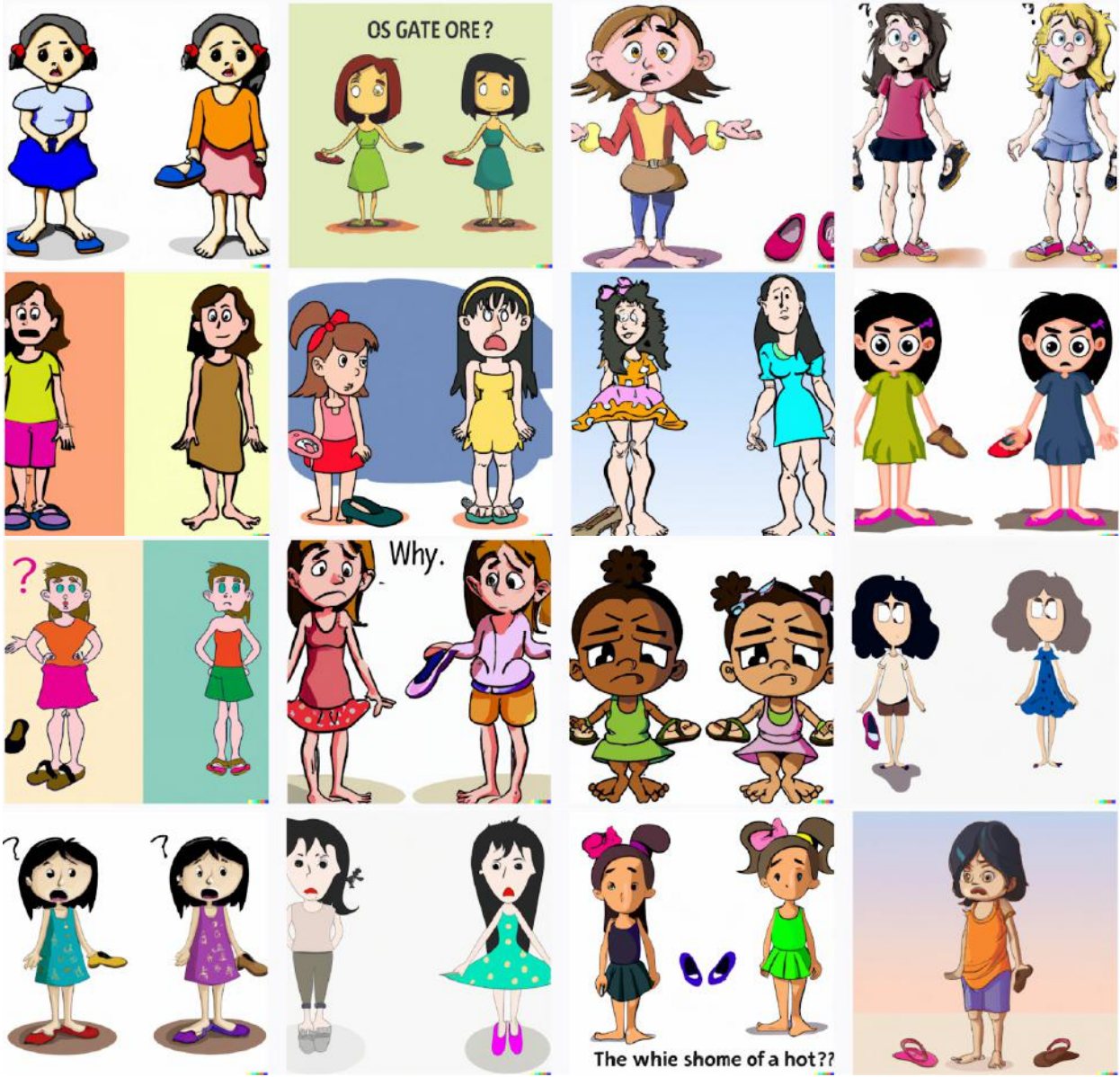

*One boy is sleeping and one boy is not sleeping*

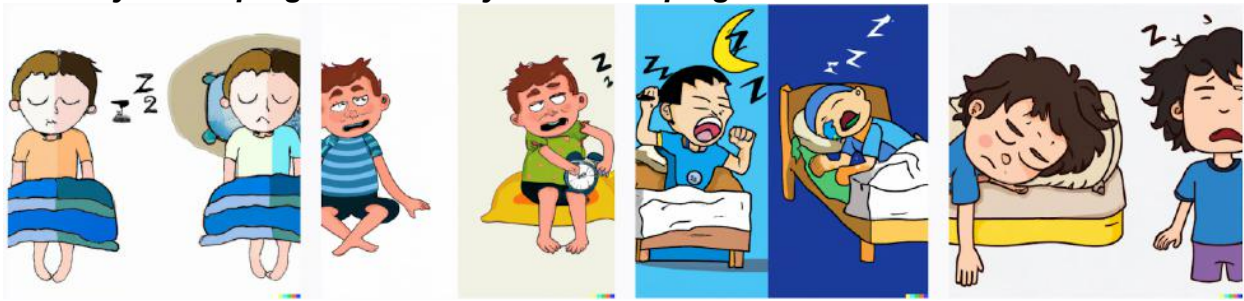

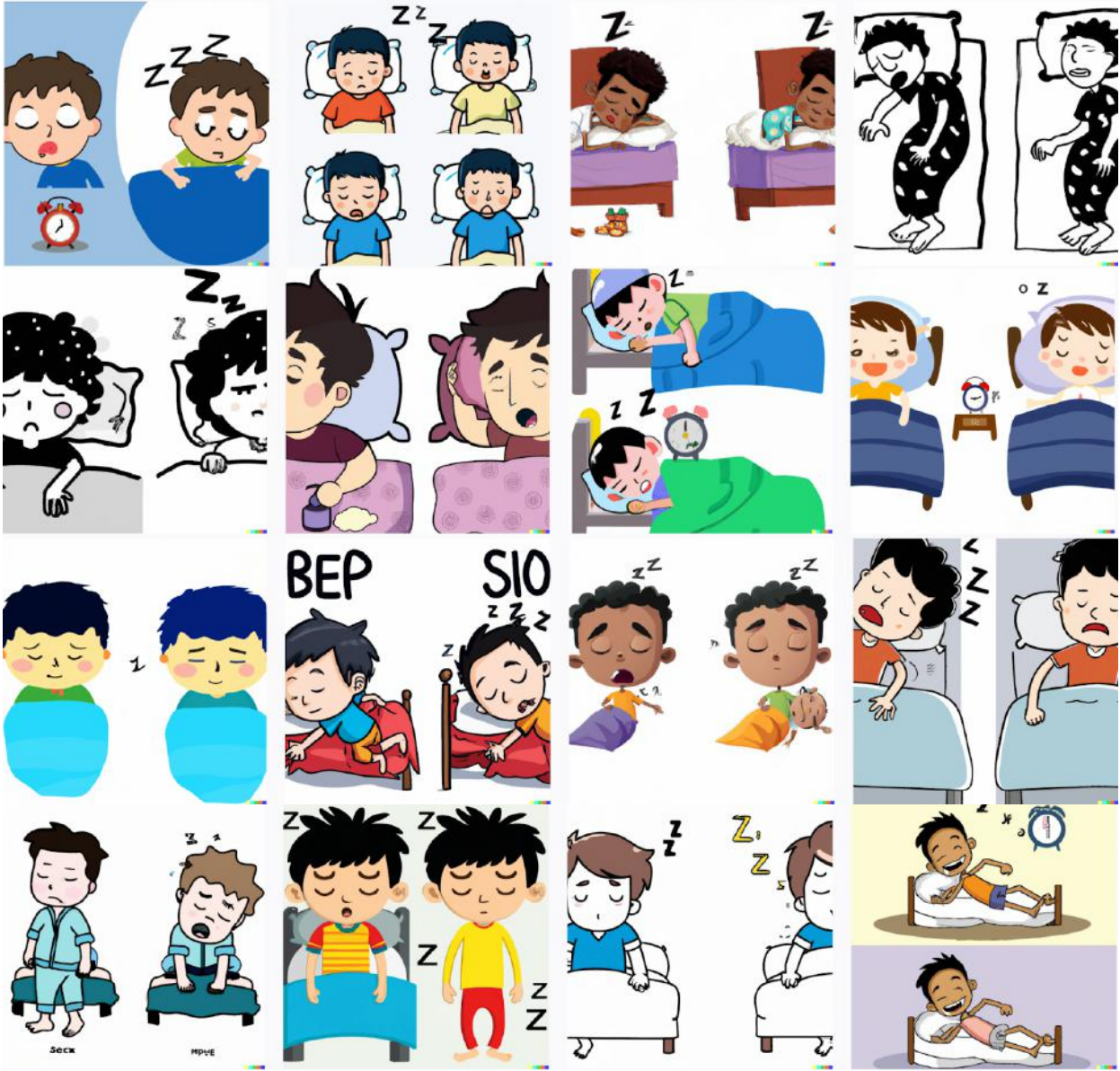

*One man has a hat and one man has no hat*

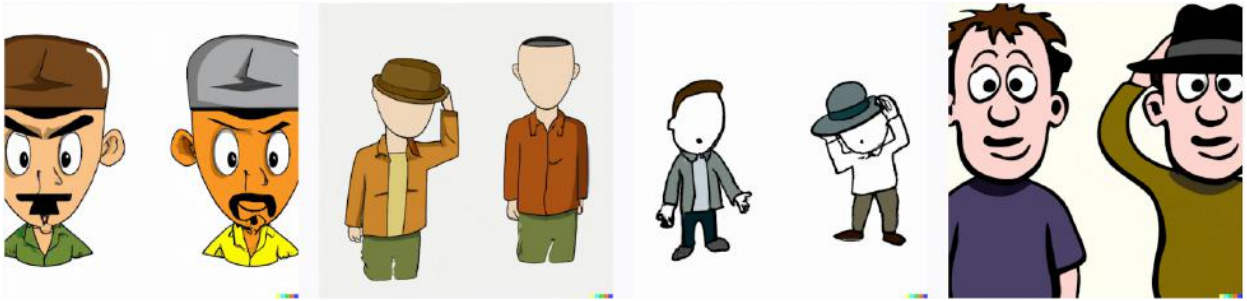

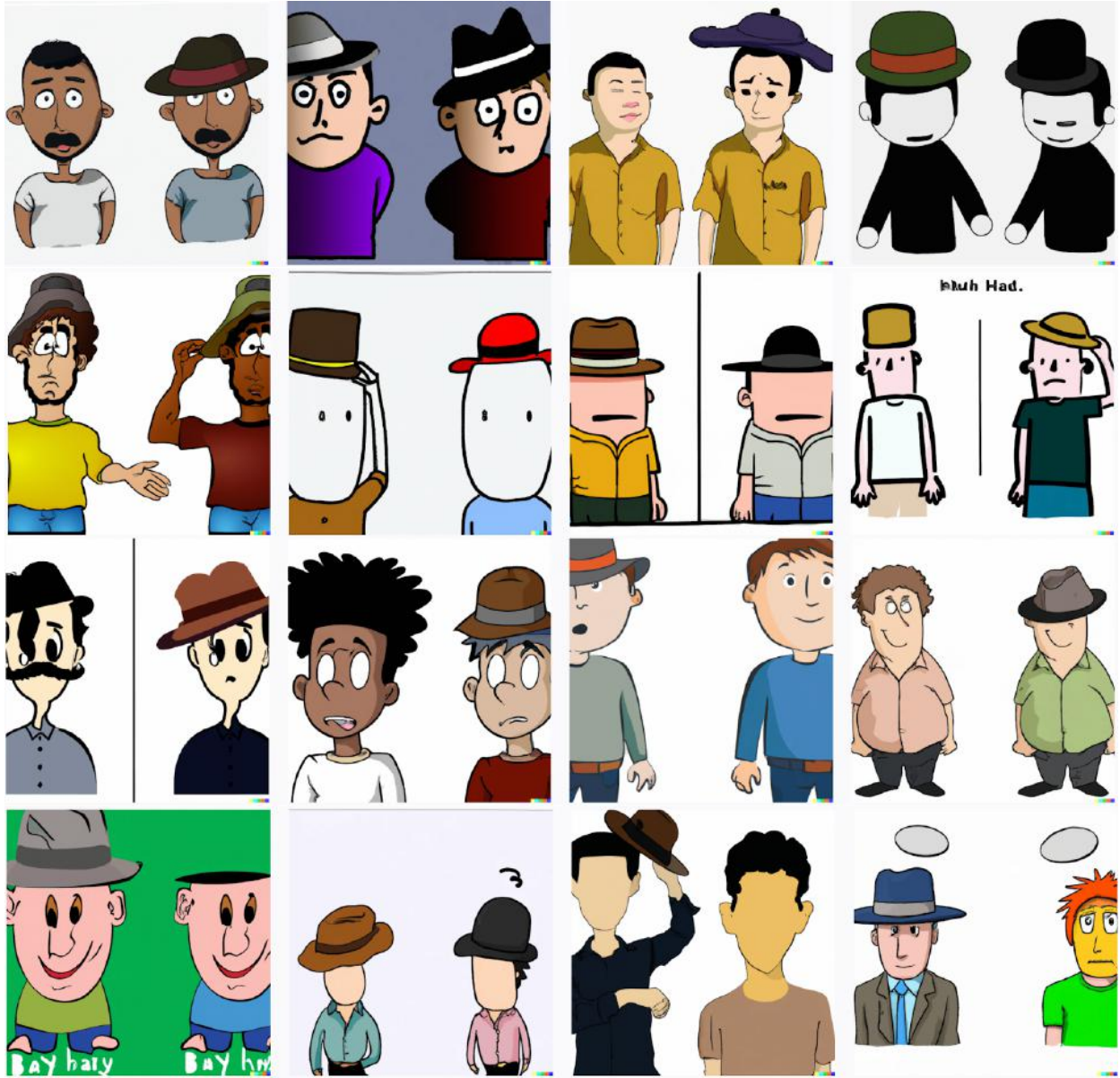

*One boy is eating food and one boy is not eating food*

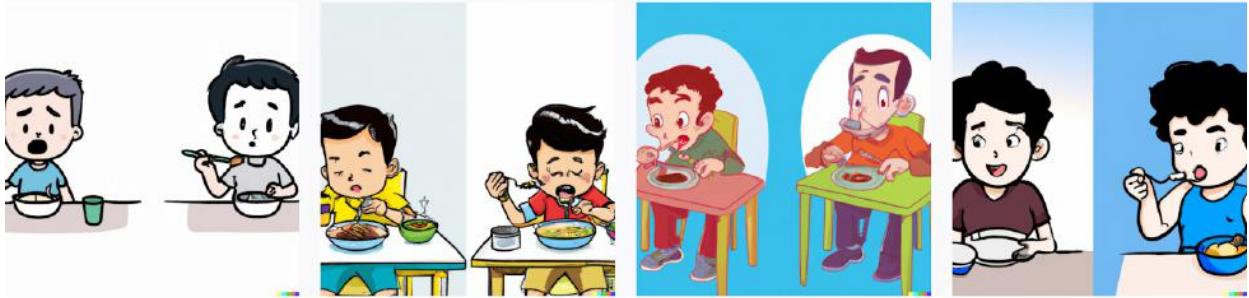

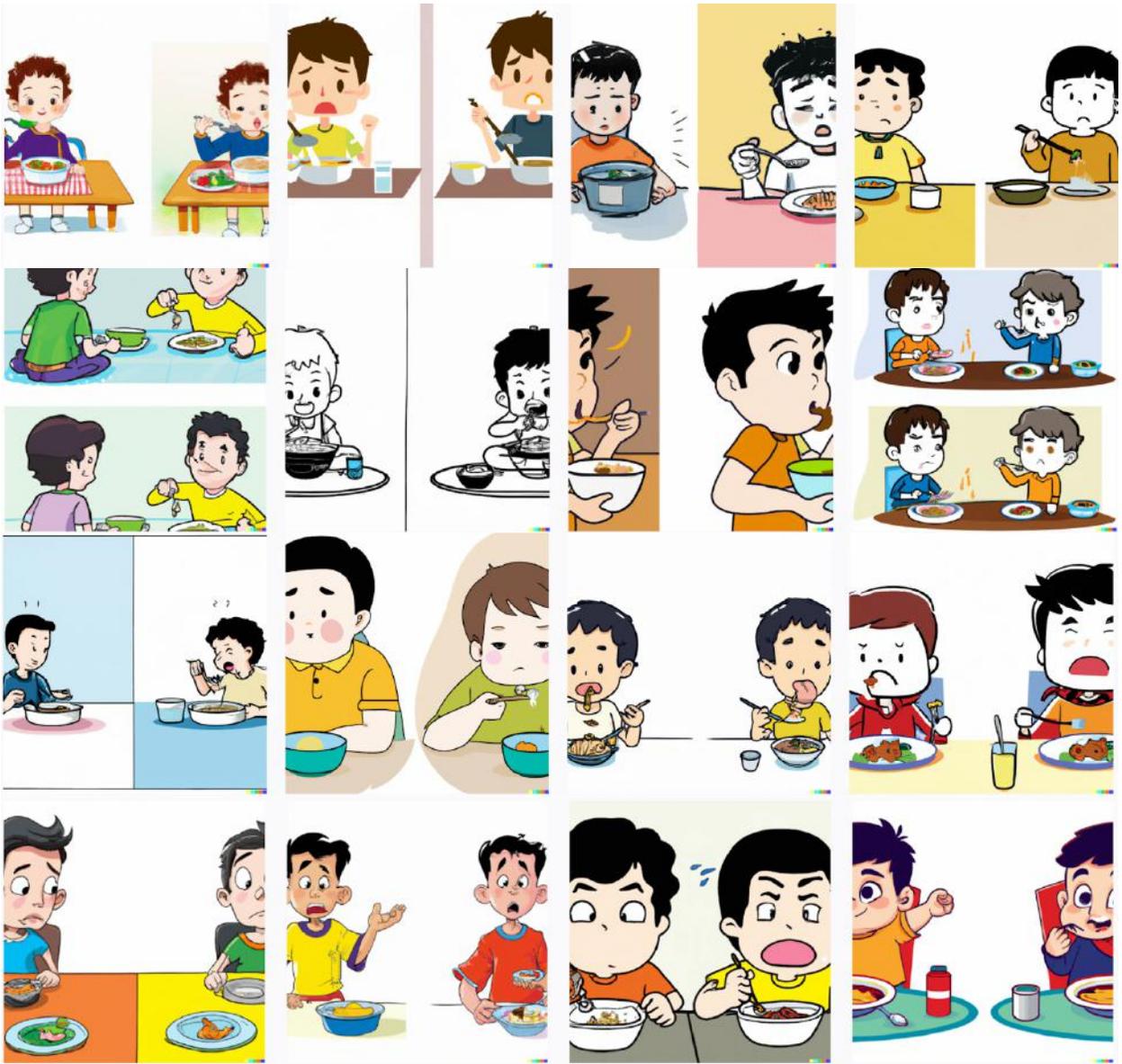

*One dog is catching a ball and one dog is not catching a ball*

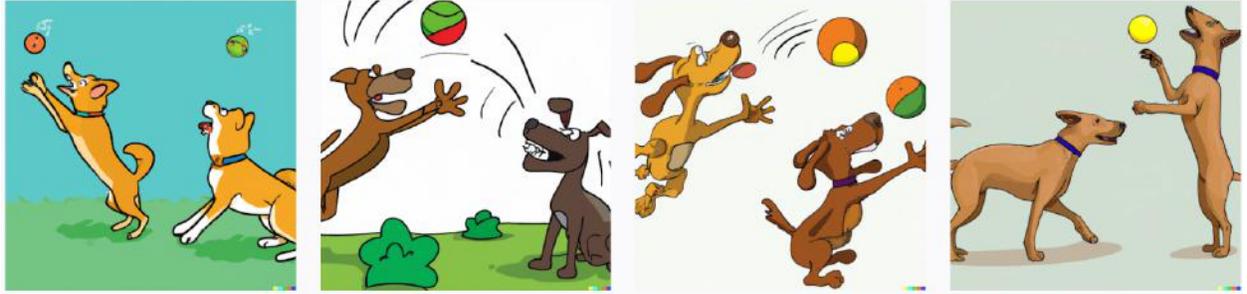

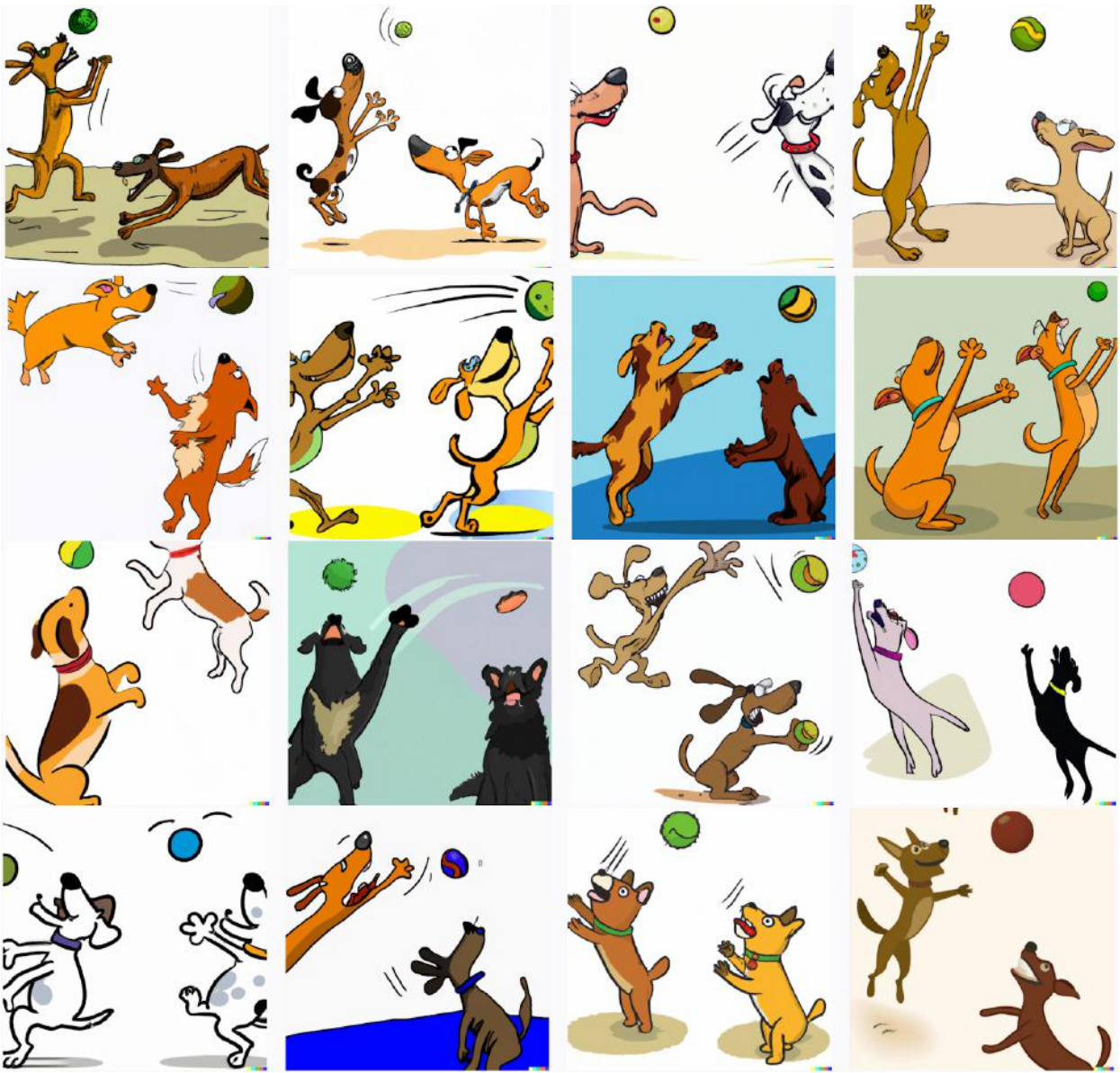

*One woman has glasses and one woman has no glasses*

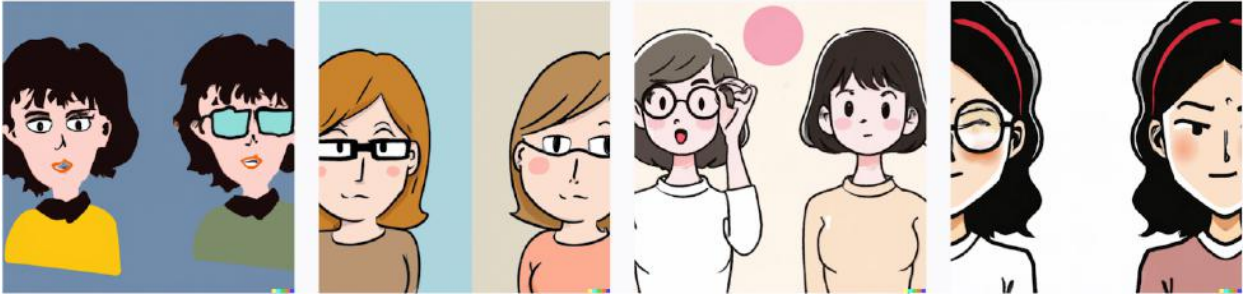

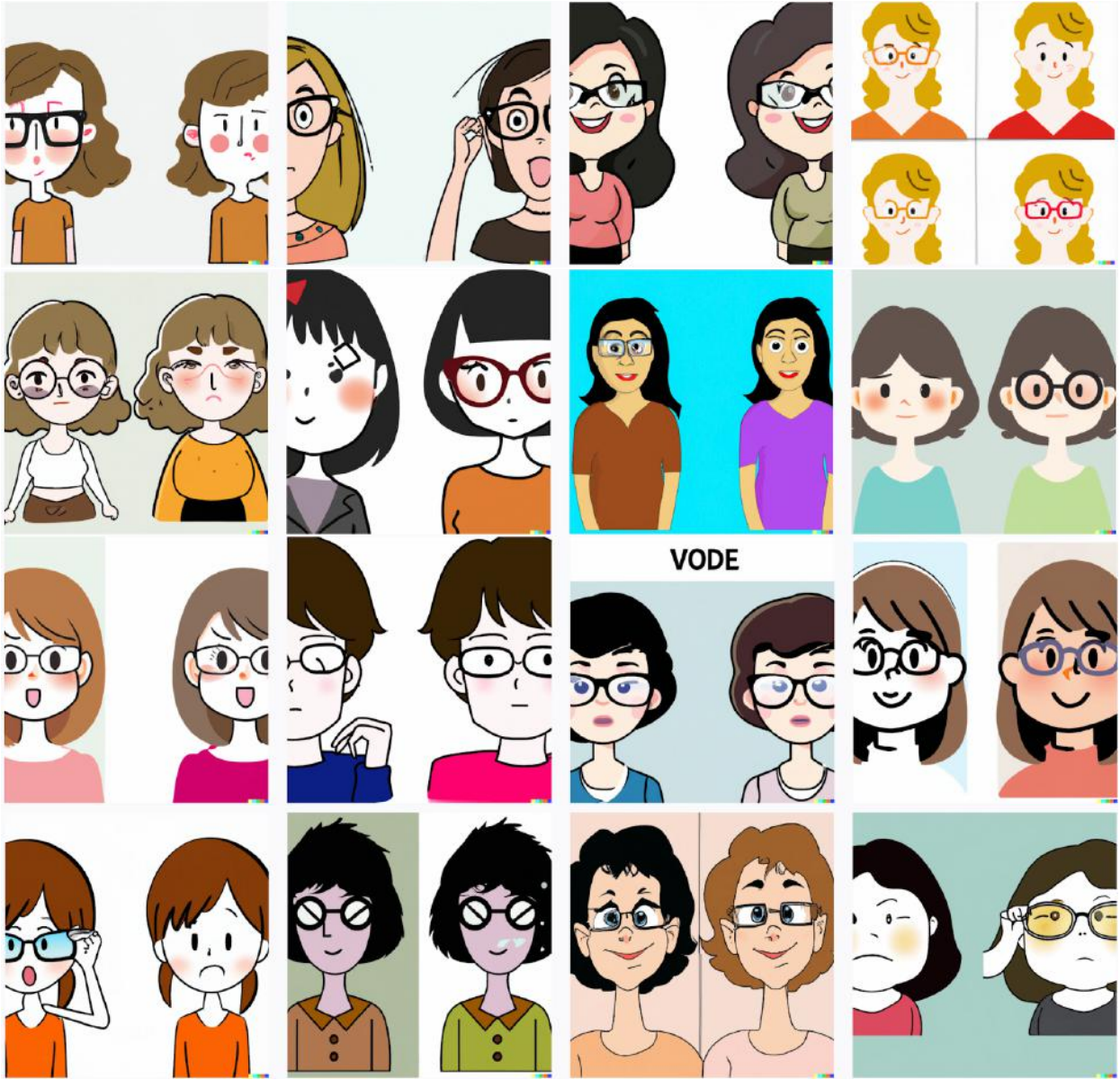

**2. Preposition examples**

*The ball is behind the dog*

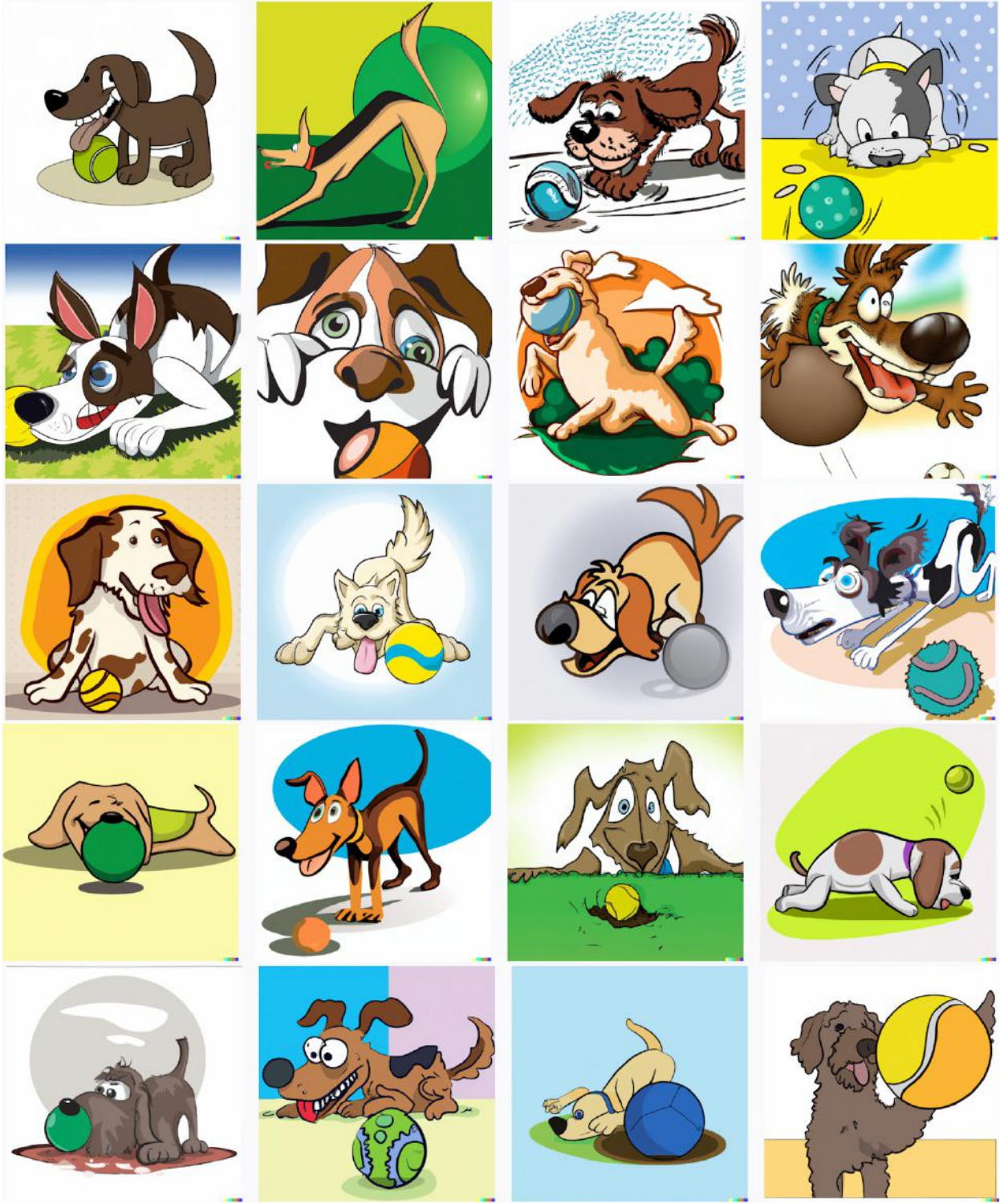

*The cup is on the book*

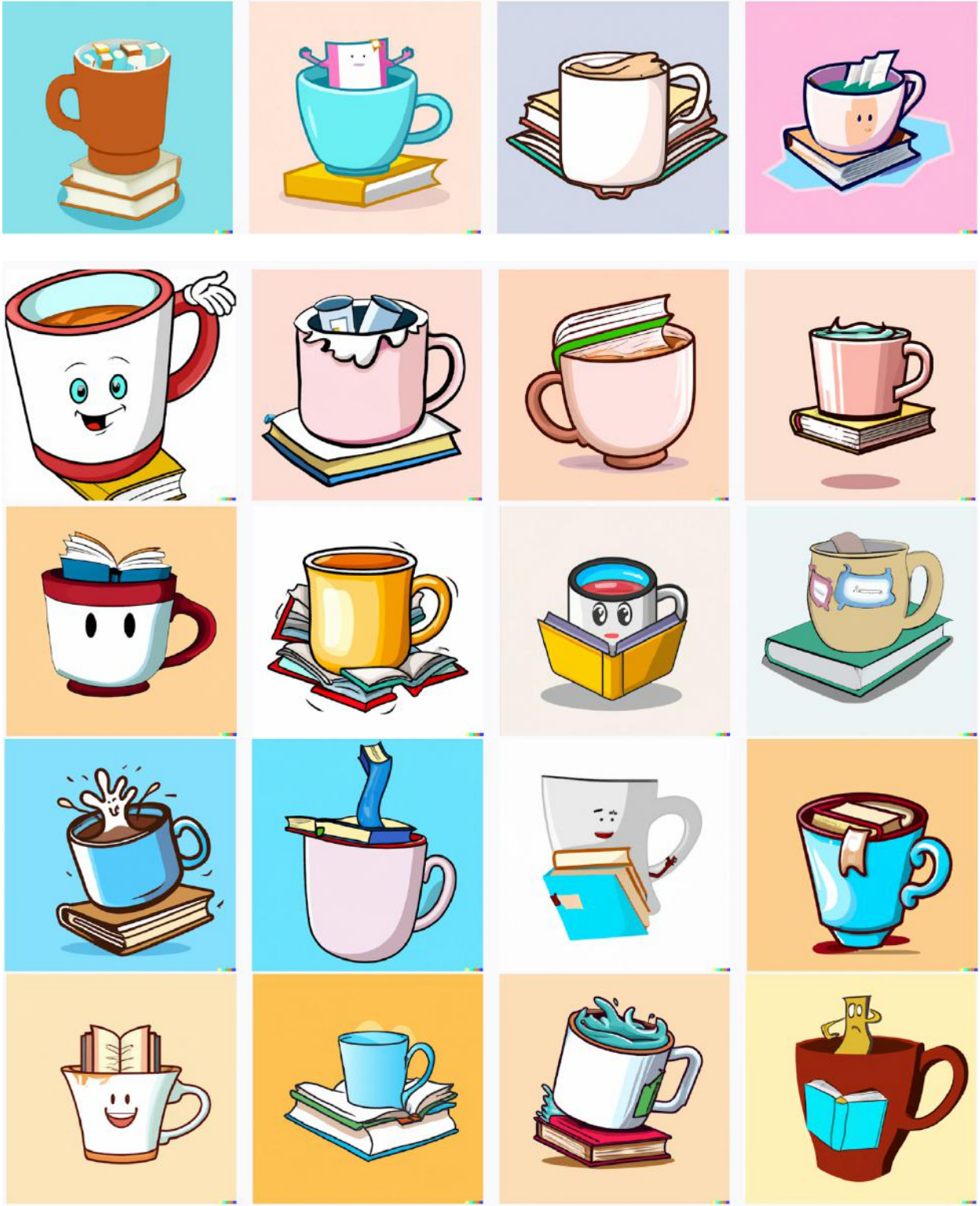

*The dog is under the chair*

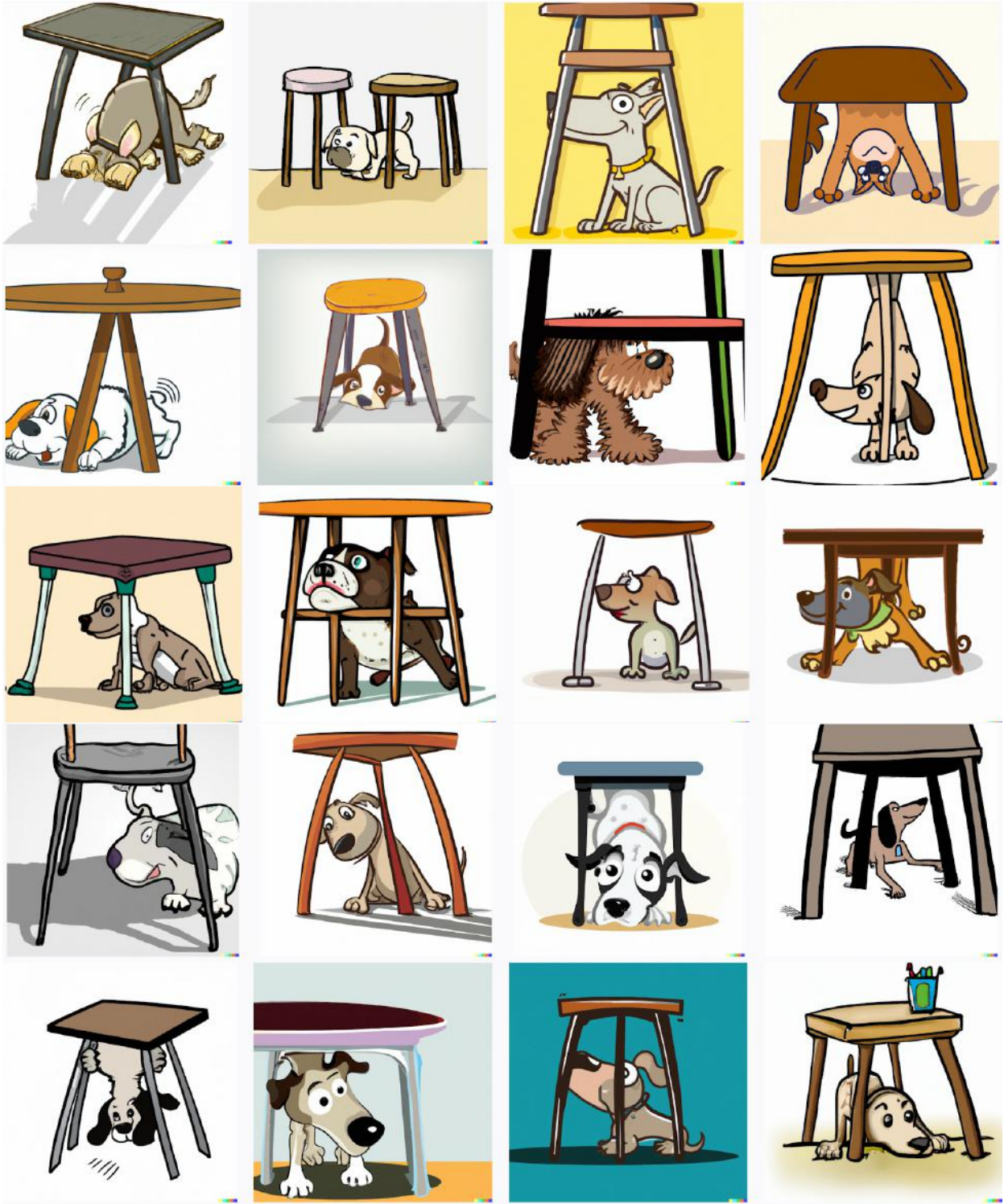

*The doll is above the present*

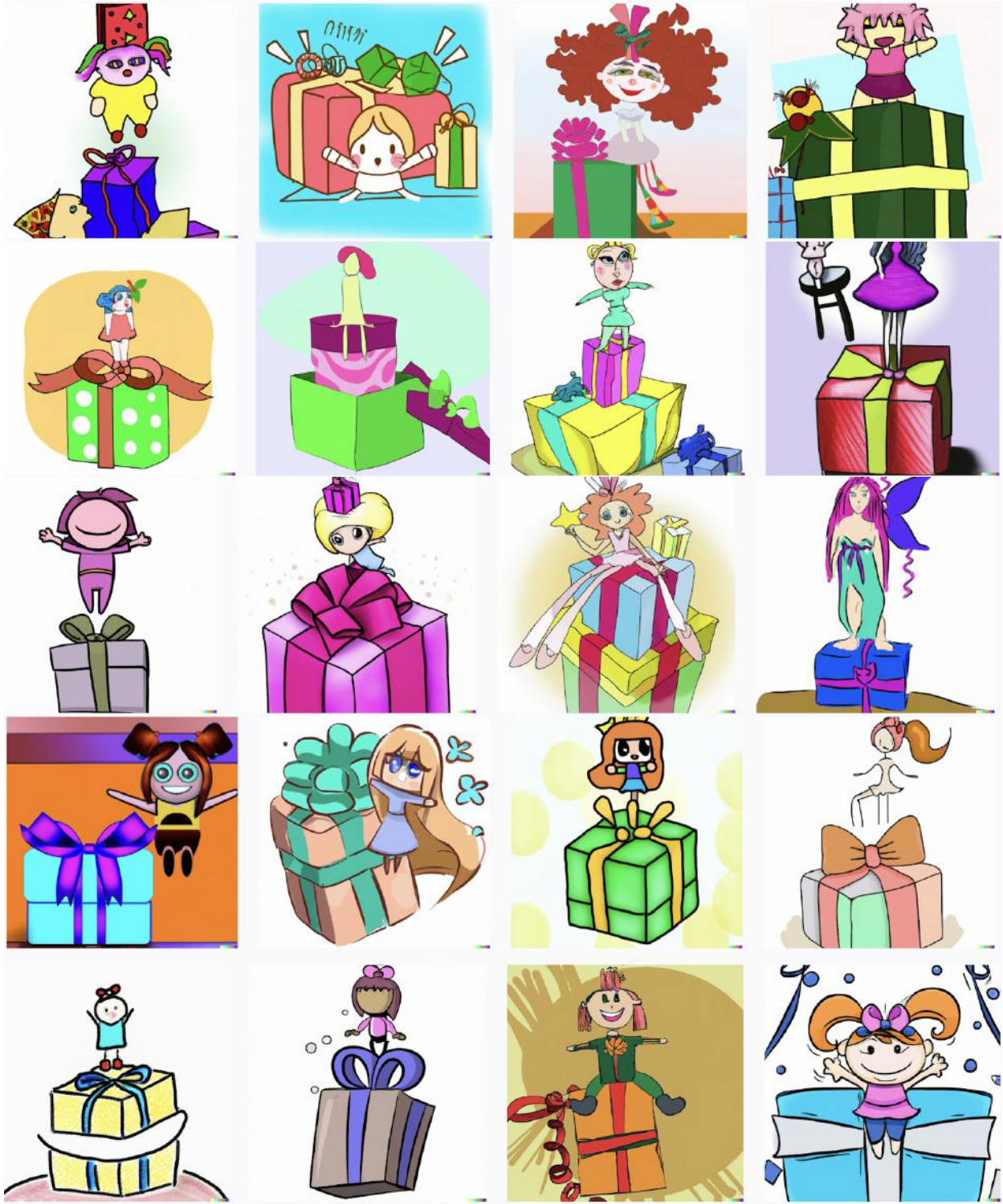

*The ball is behind the pail*

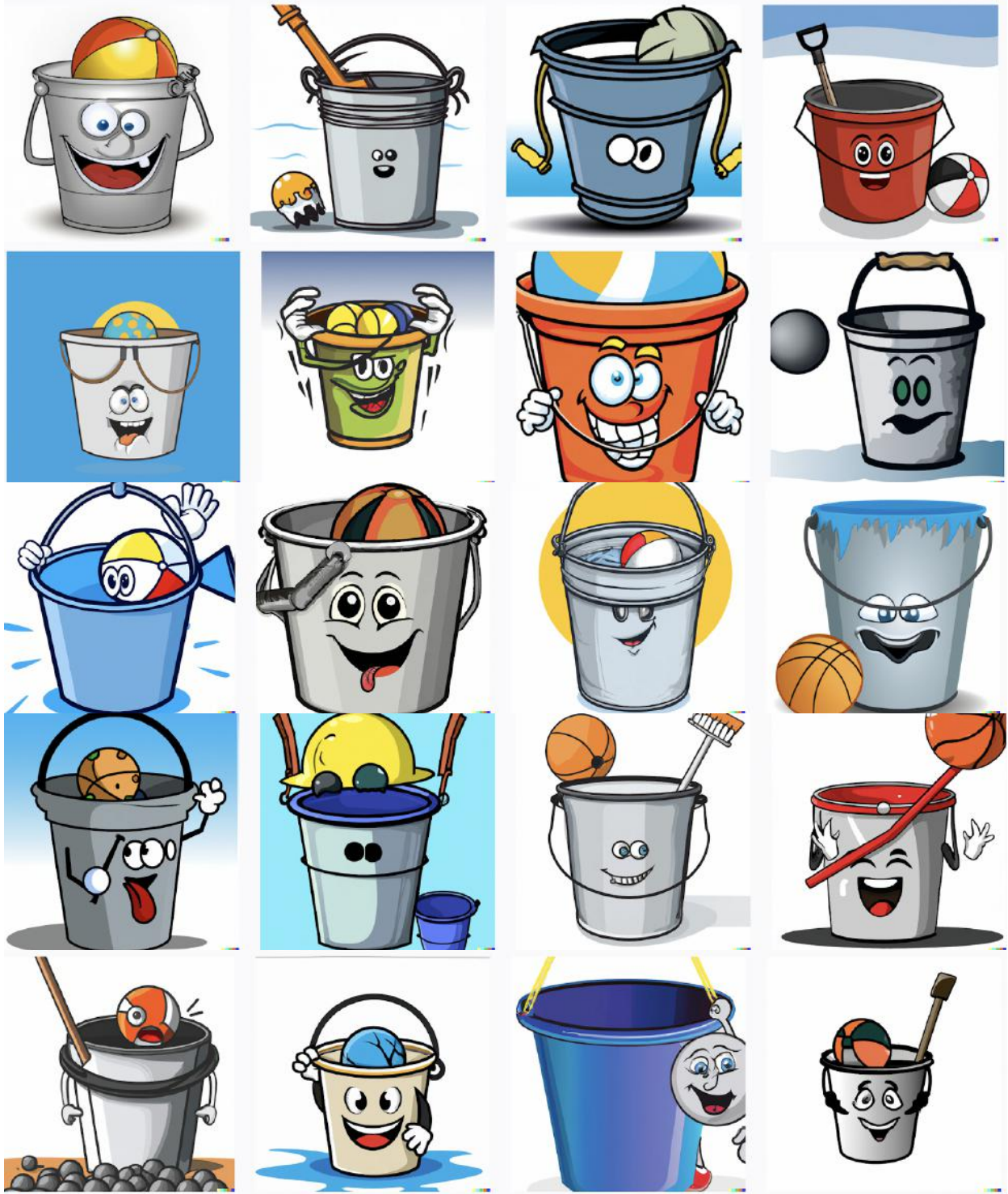

*The firefighters are between the chairs*

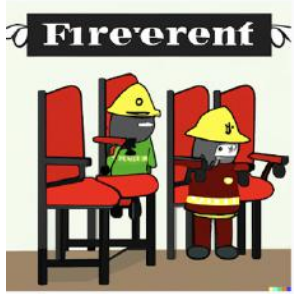 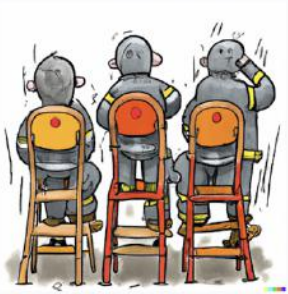 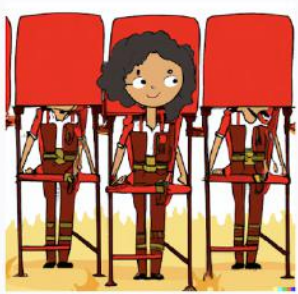 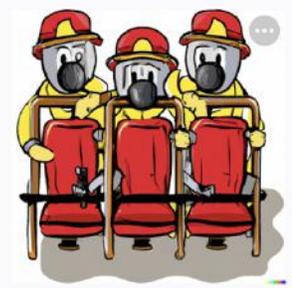
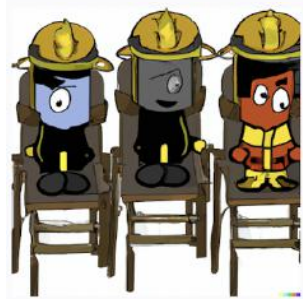 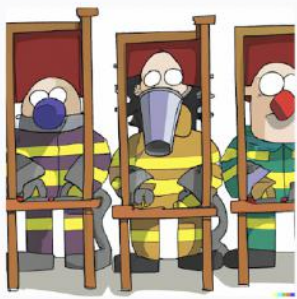 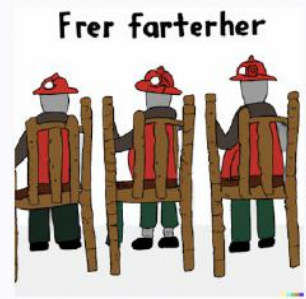 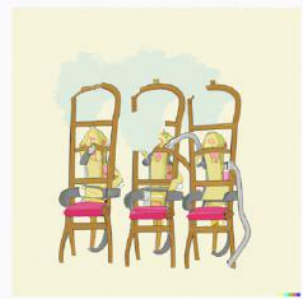
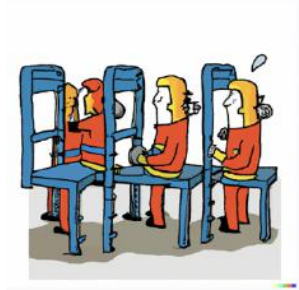 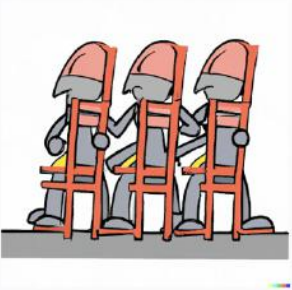 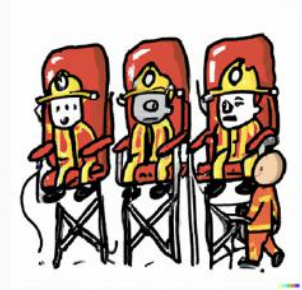 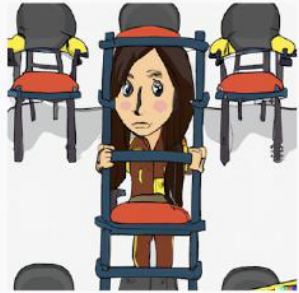
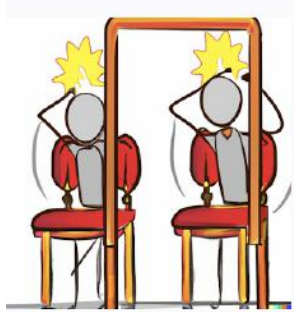 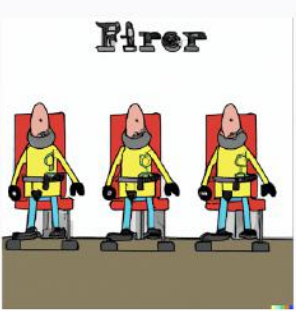 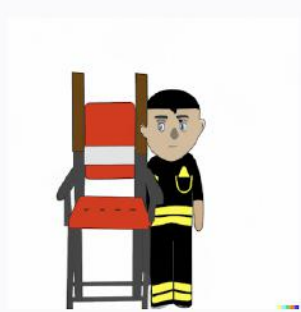 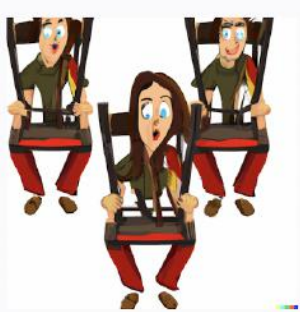
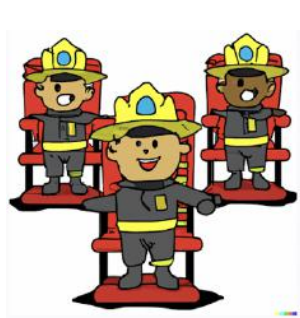 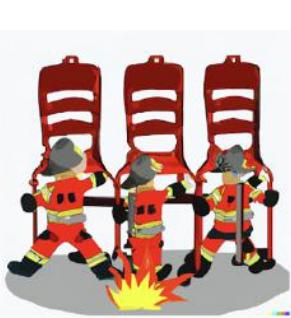 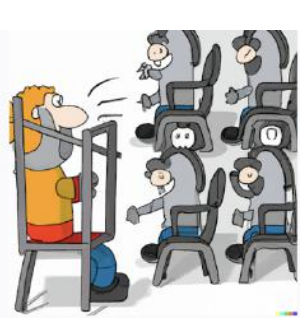 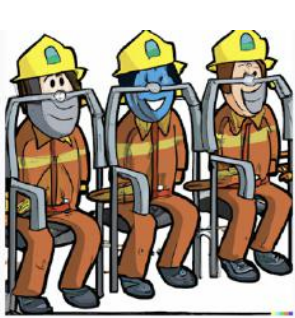

*The umbrella is below the swing*

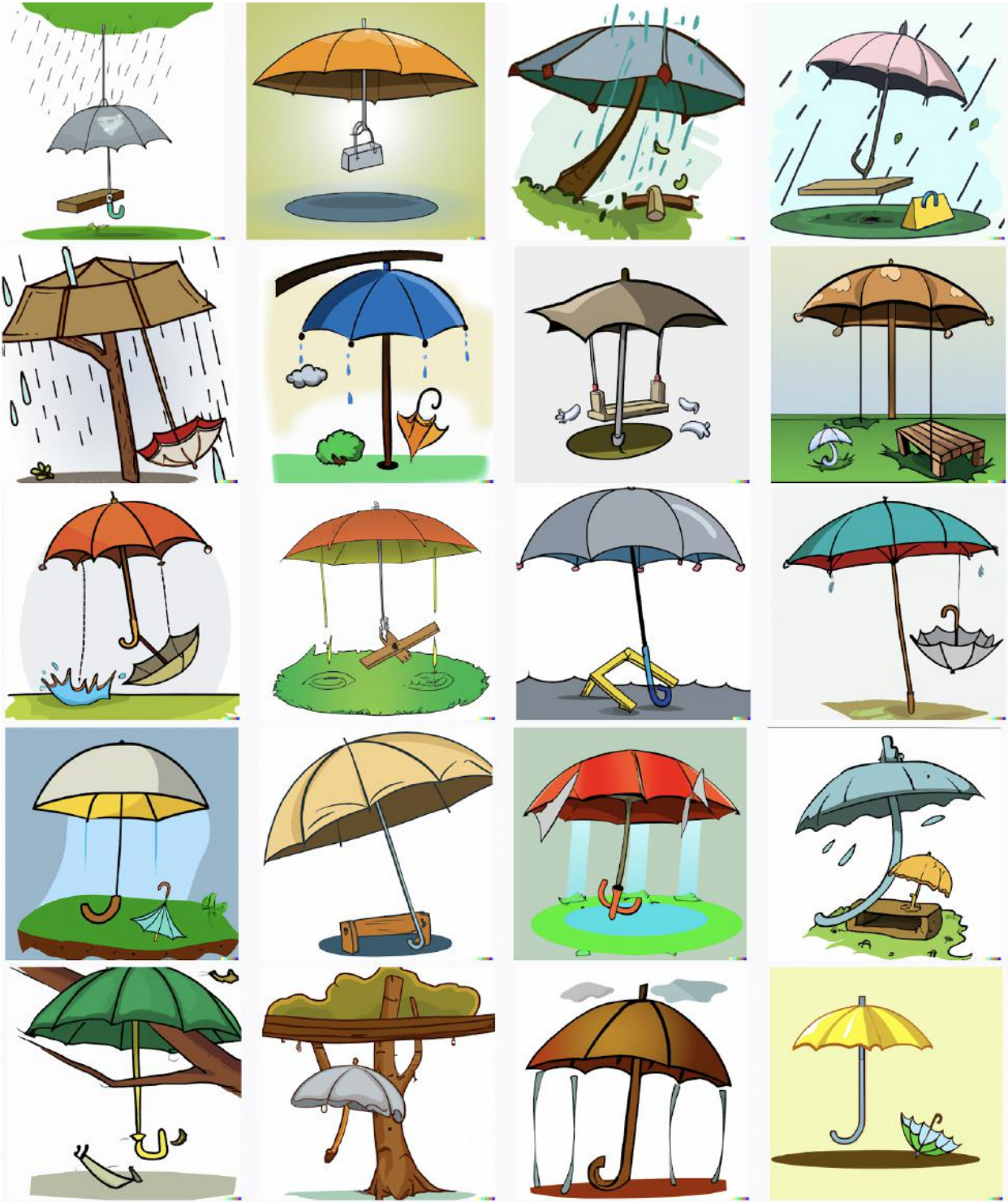

*The dog is behind a black table*

*The kitten is in a cup with a yellow ribbon*

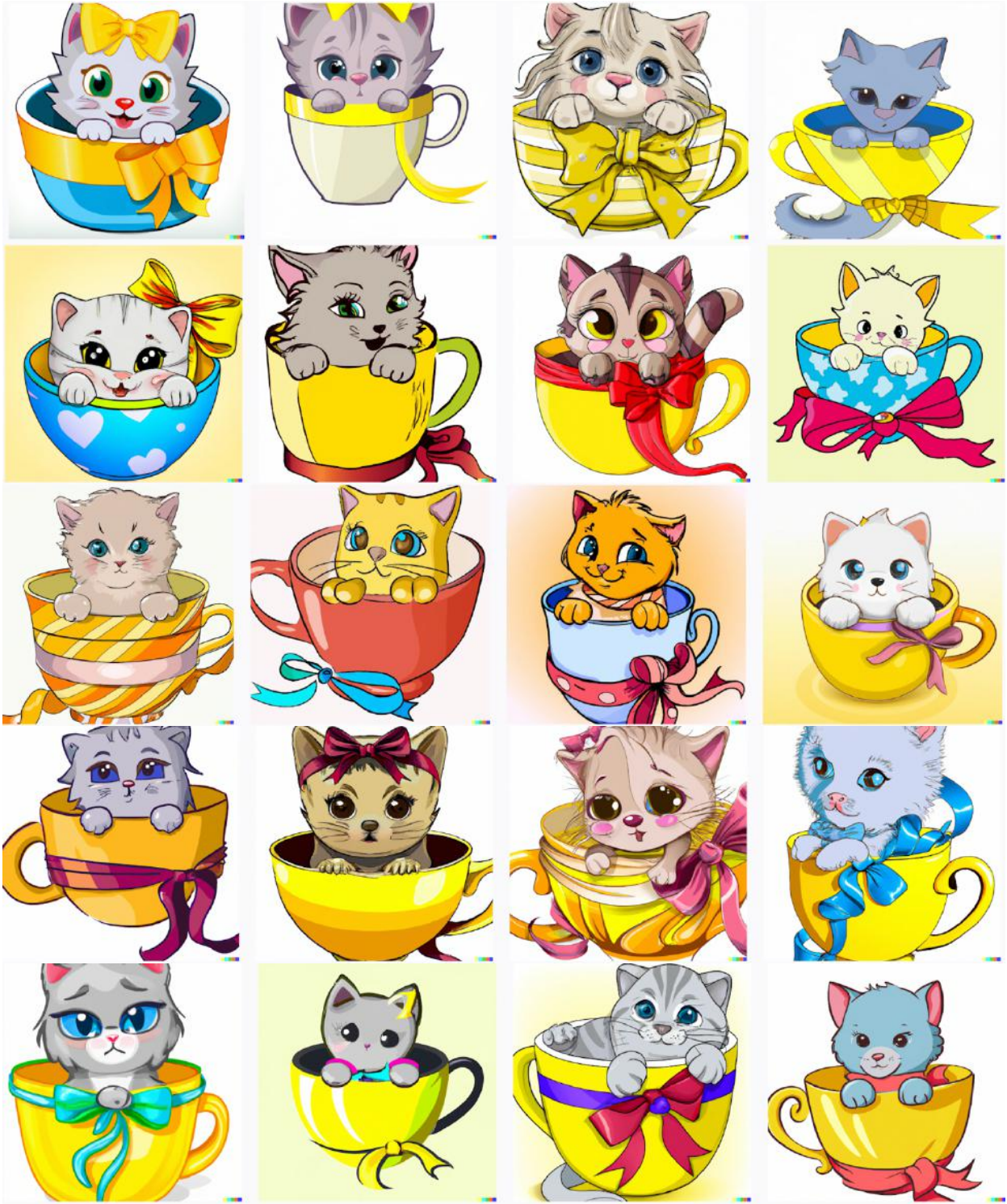

*The girl is behind a car in a white garage*

3. **Transitives**

*The car bumps the truck*

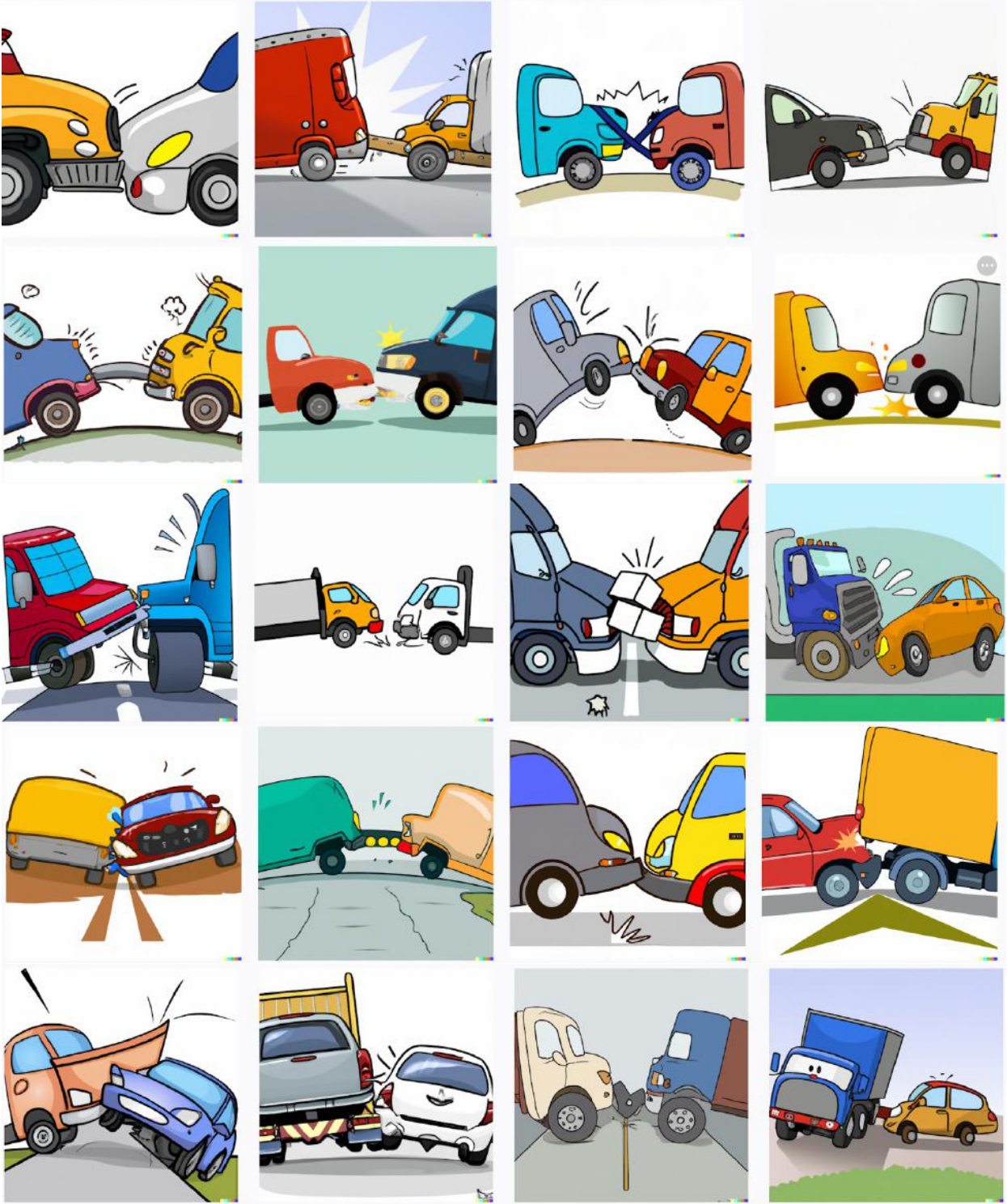

*The girl sprays the boy*

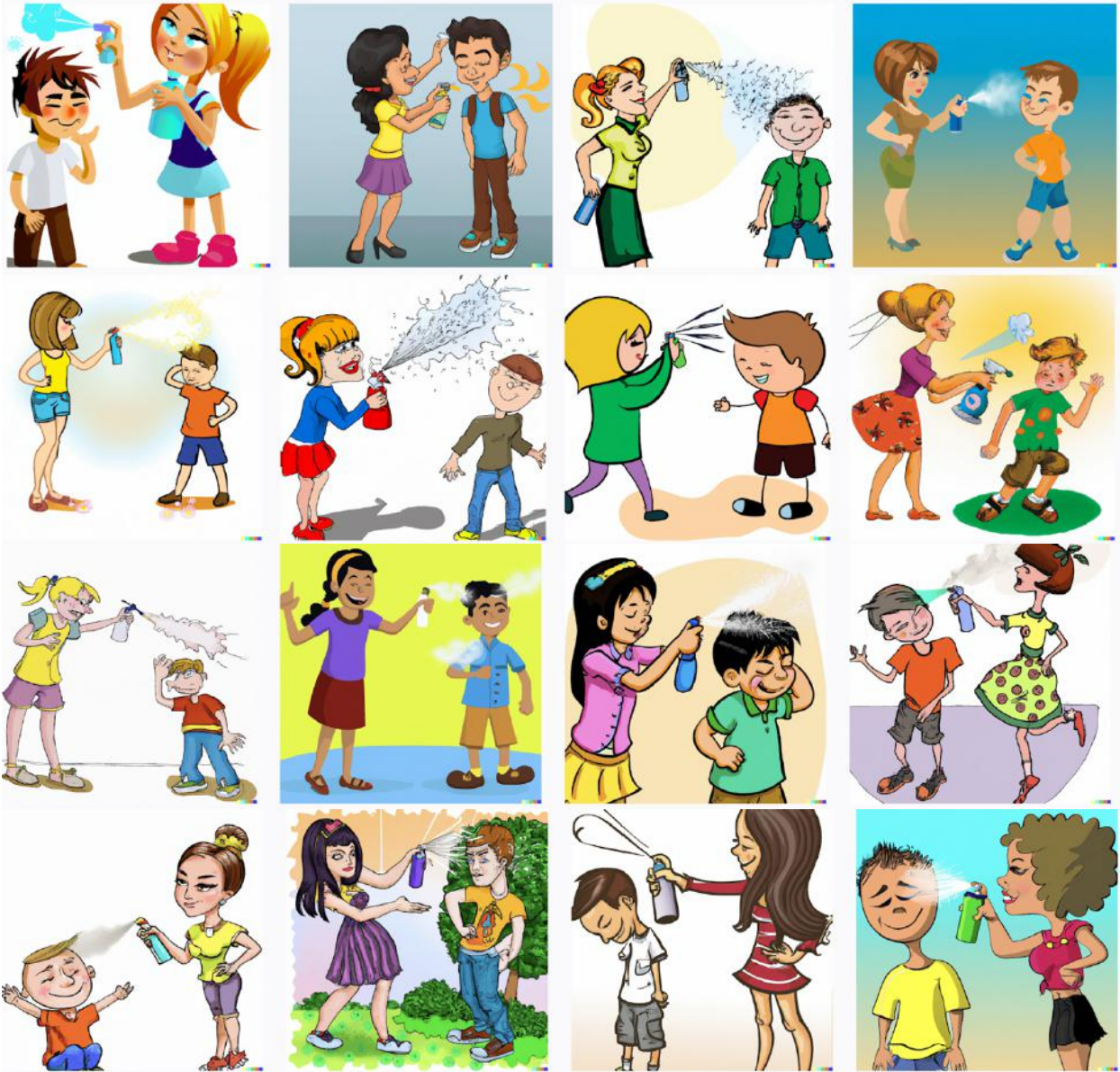

*The horse paints the elephant*

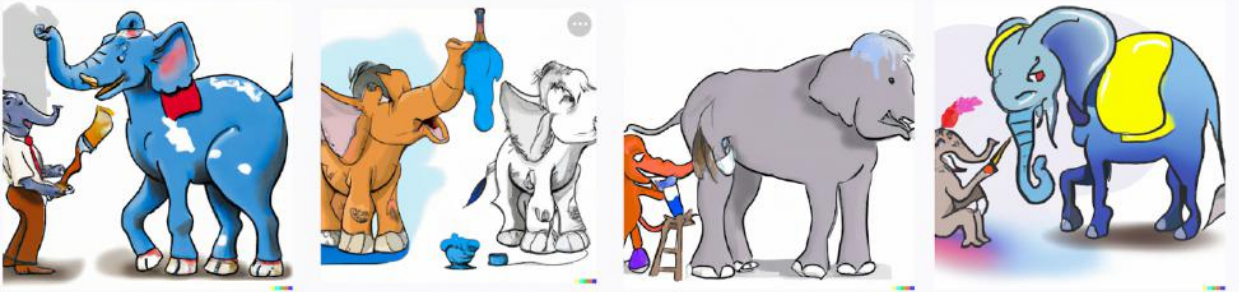

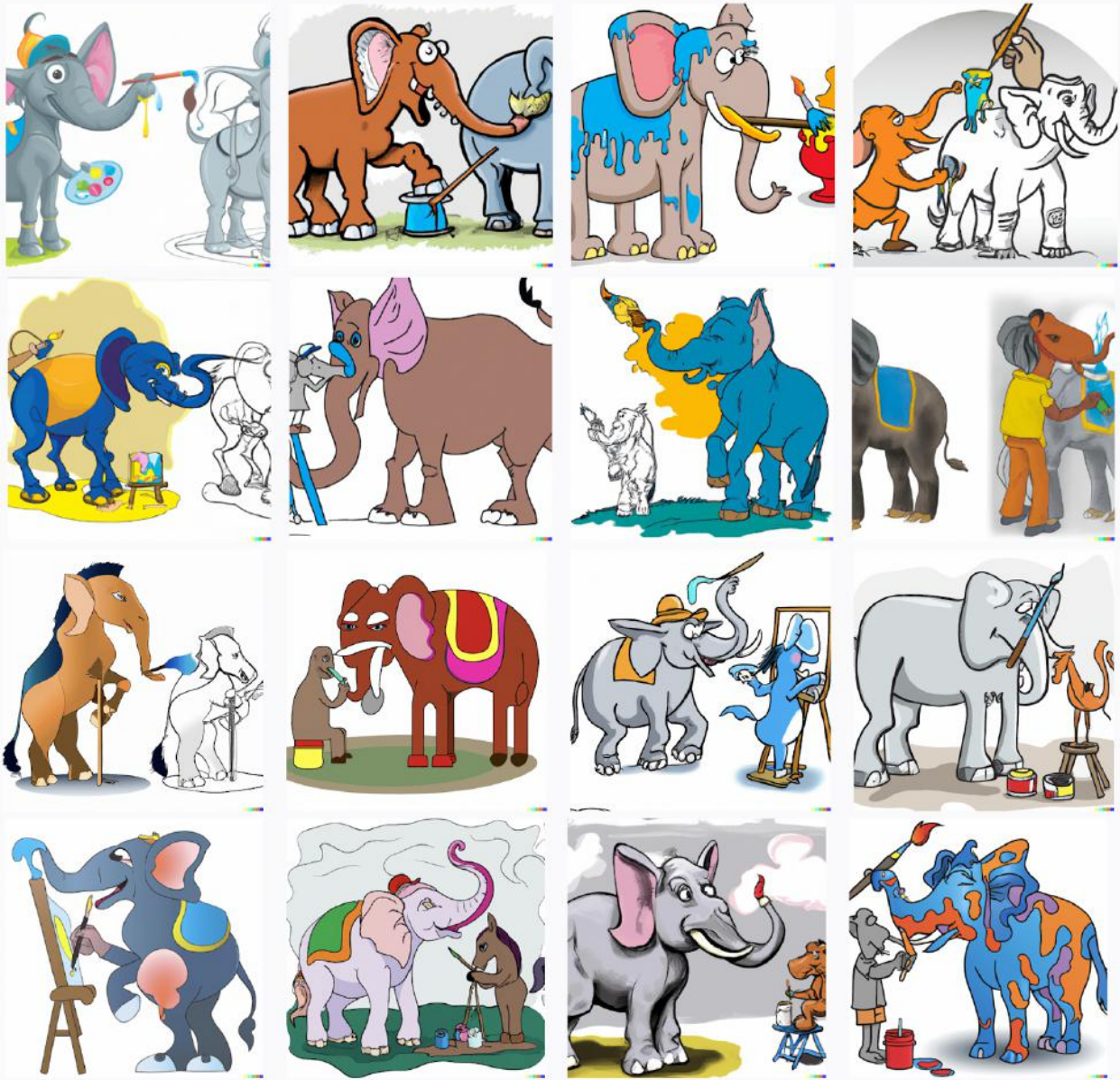

*The duck is pushing the bunny*

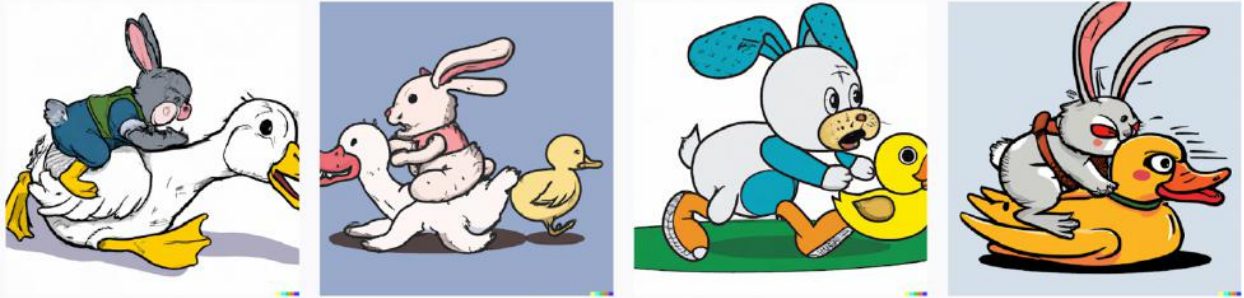

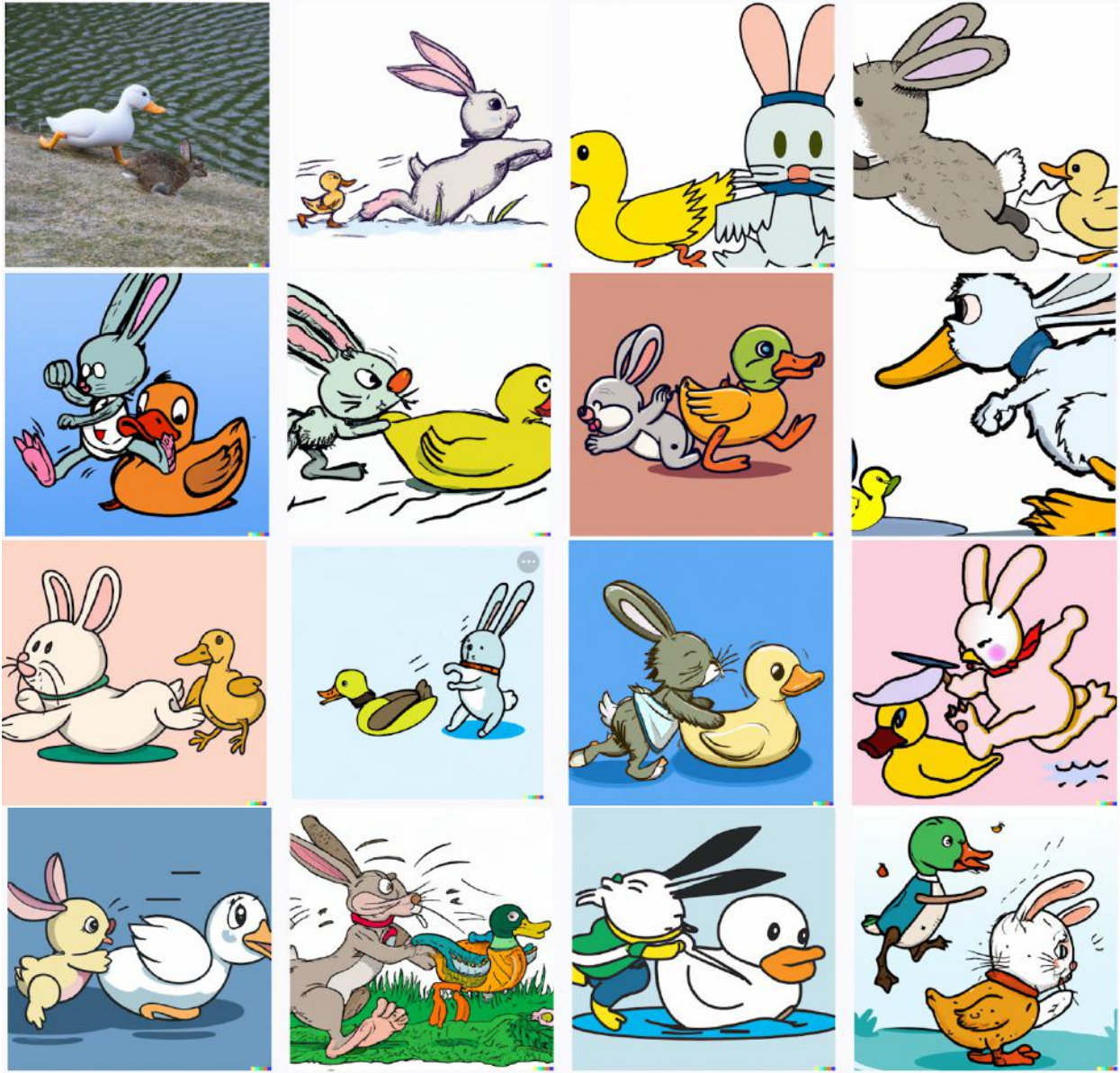

*The dog is pushing the pig*

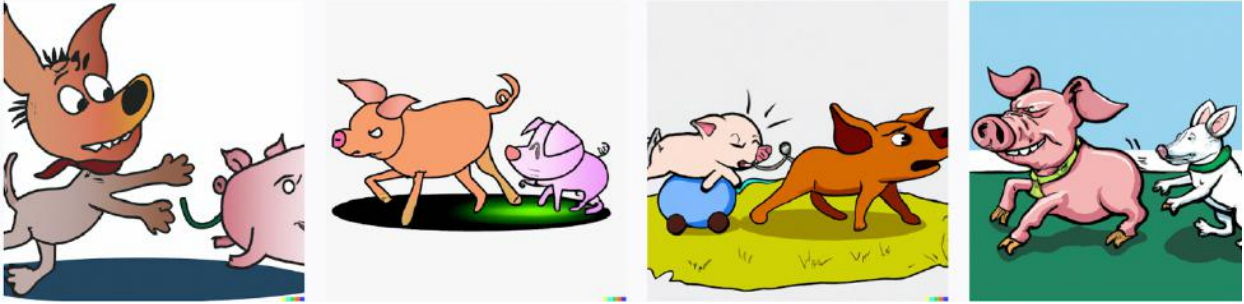

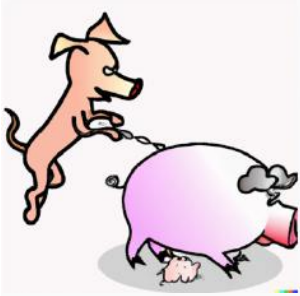 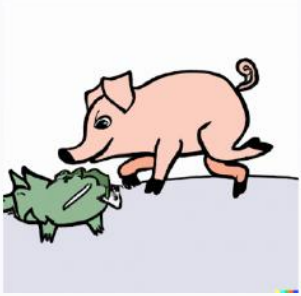 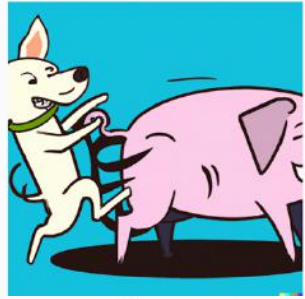 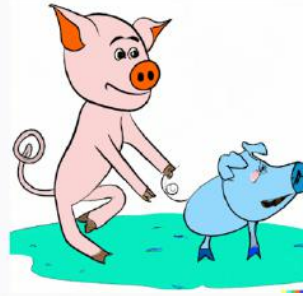
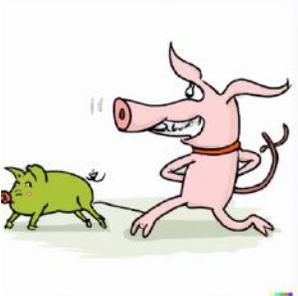 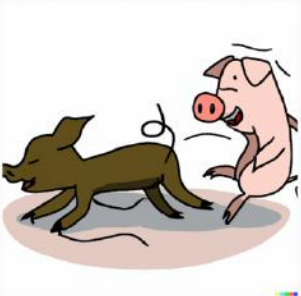 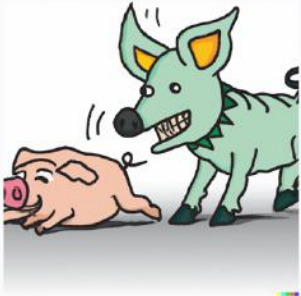 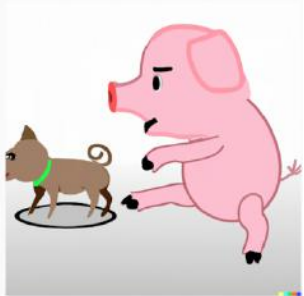
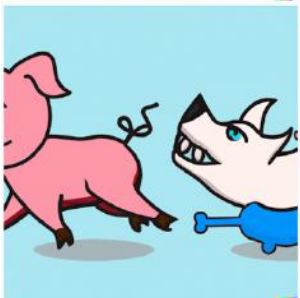 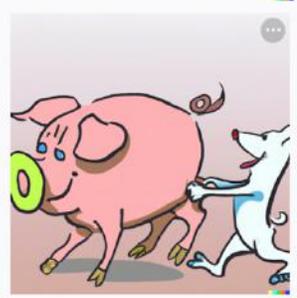 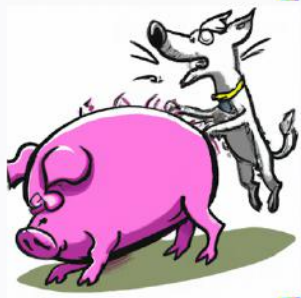 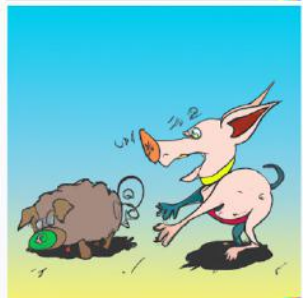
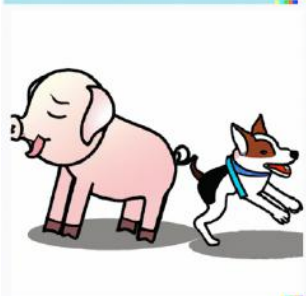 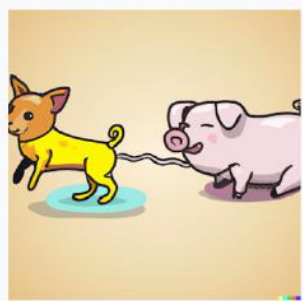 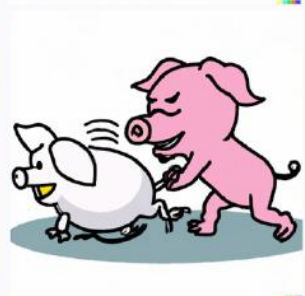 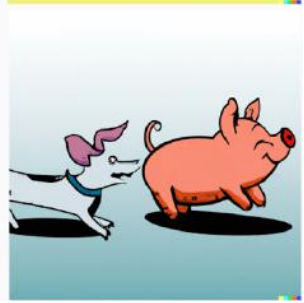

*The horse is pushing the cow*

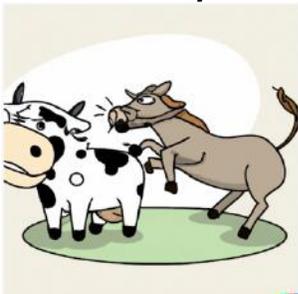 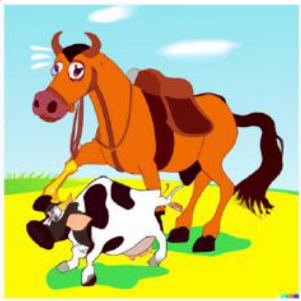 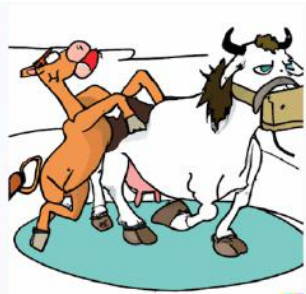 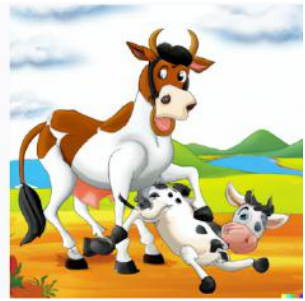

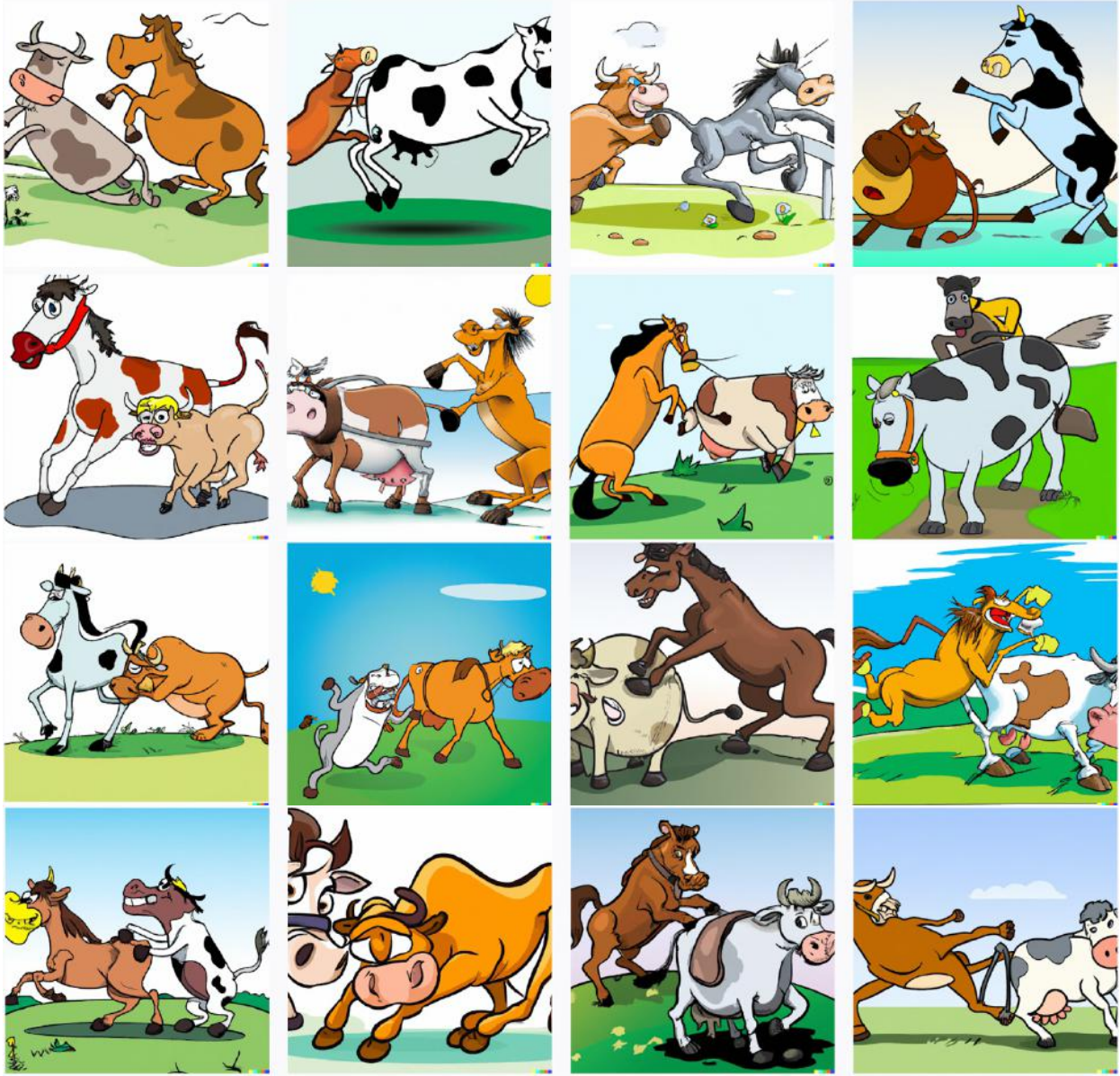

*The bear is pushing the lion*

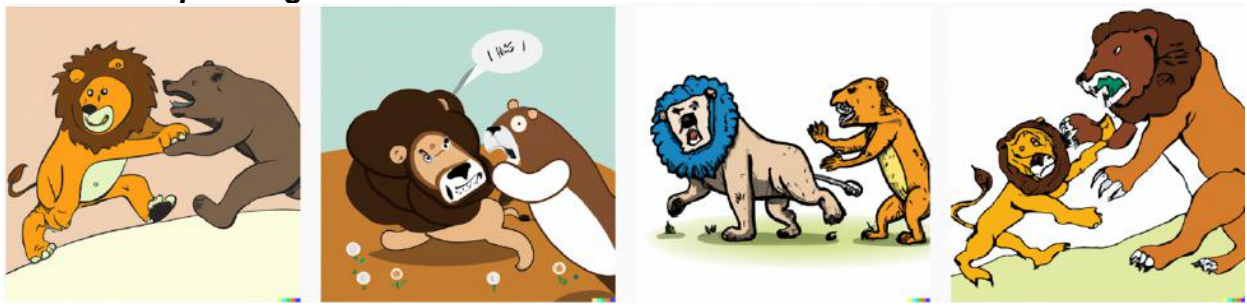

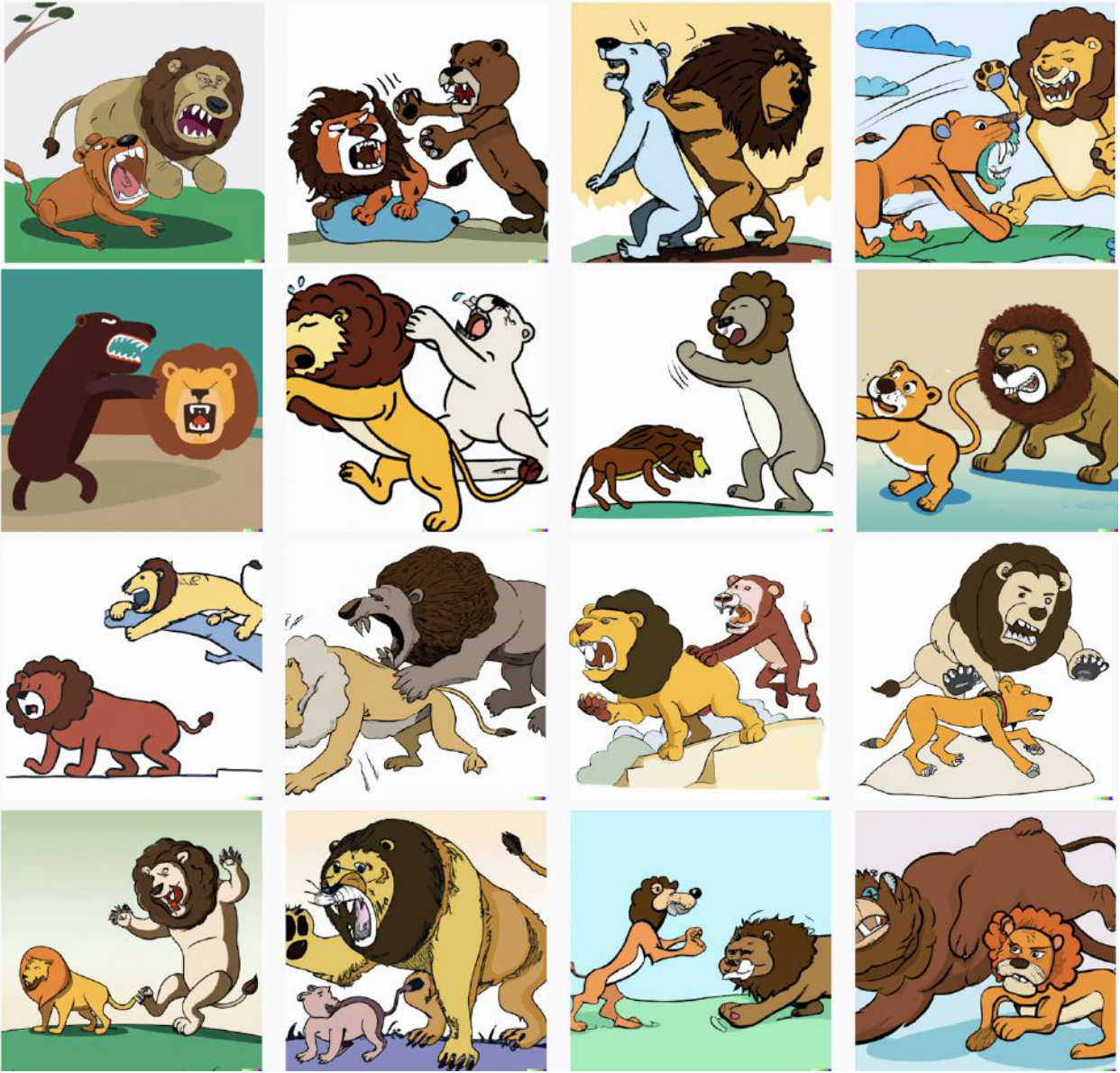

*The elephant is kicking the tiger*

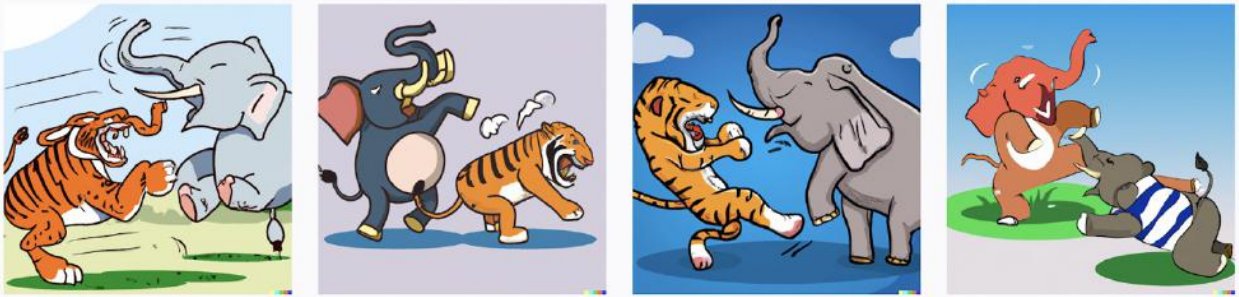

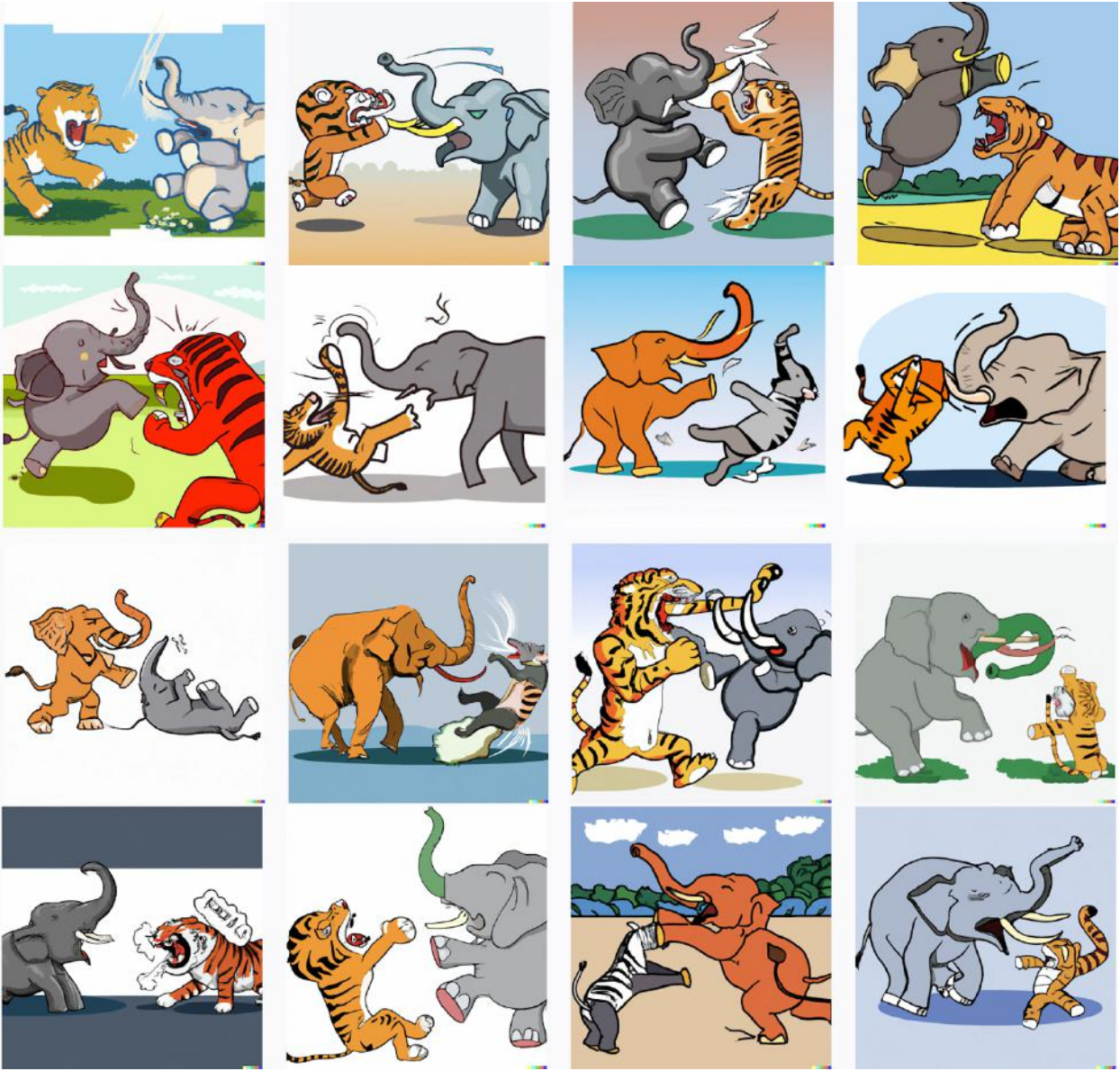

*The spider is kicking the frog*

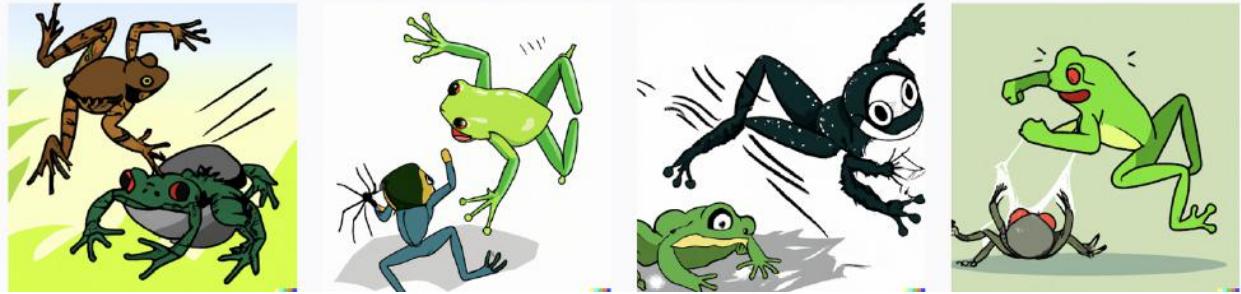

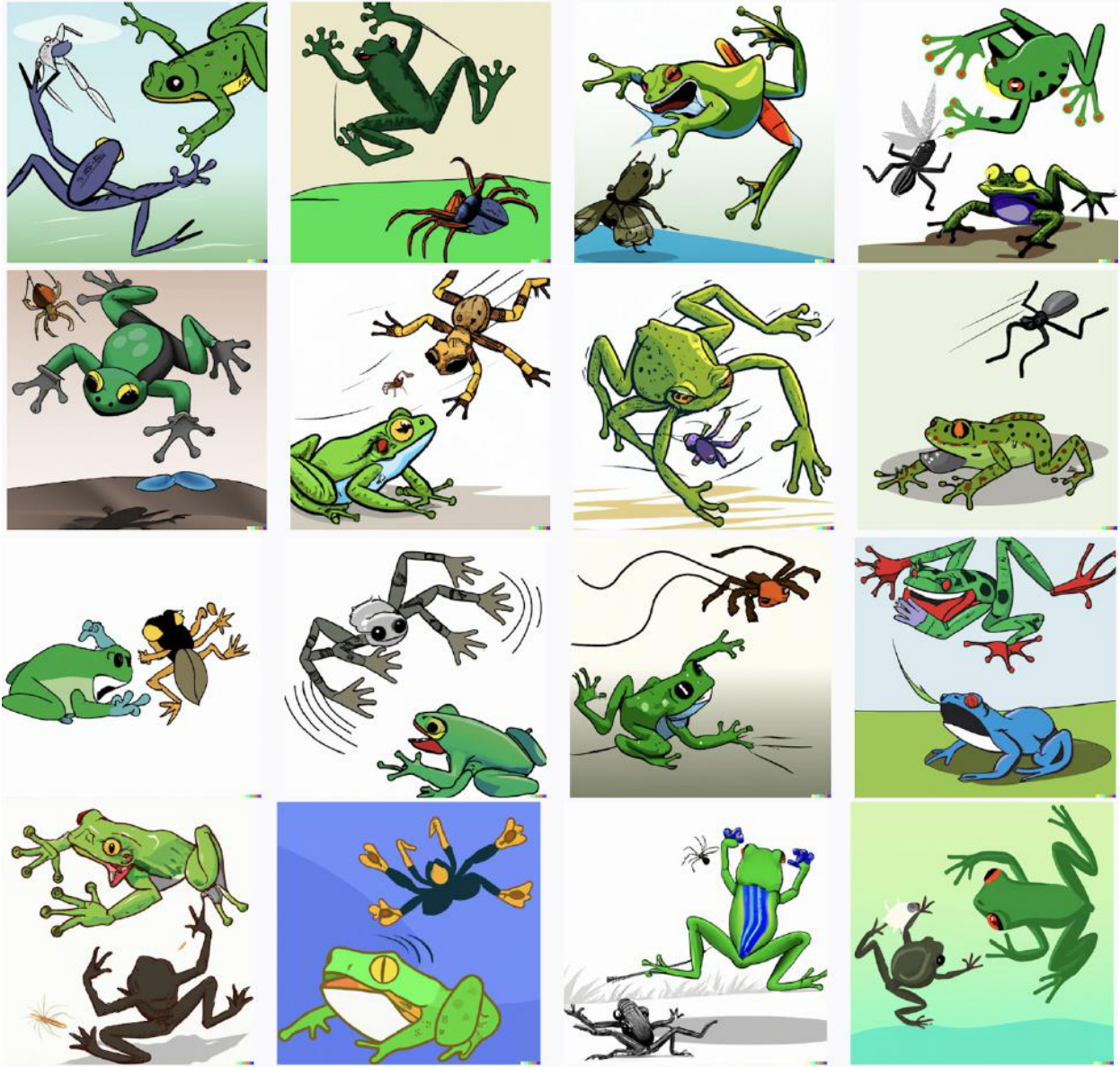

*The chicken is kicking the mouse*

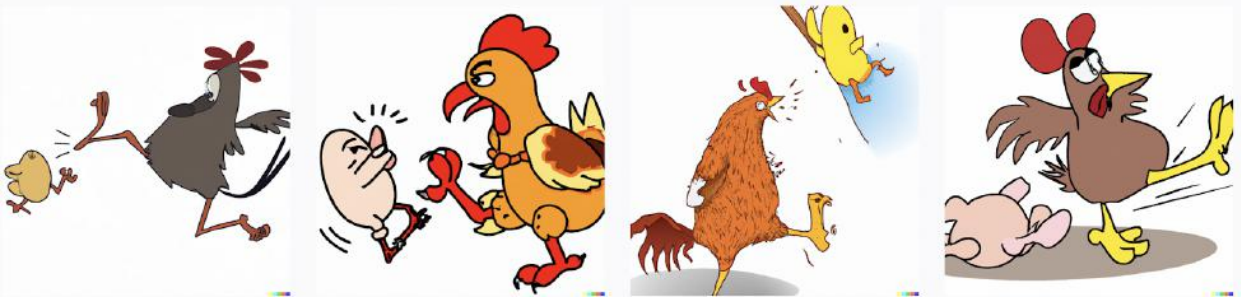

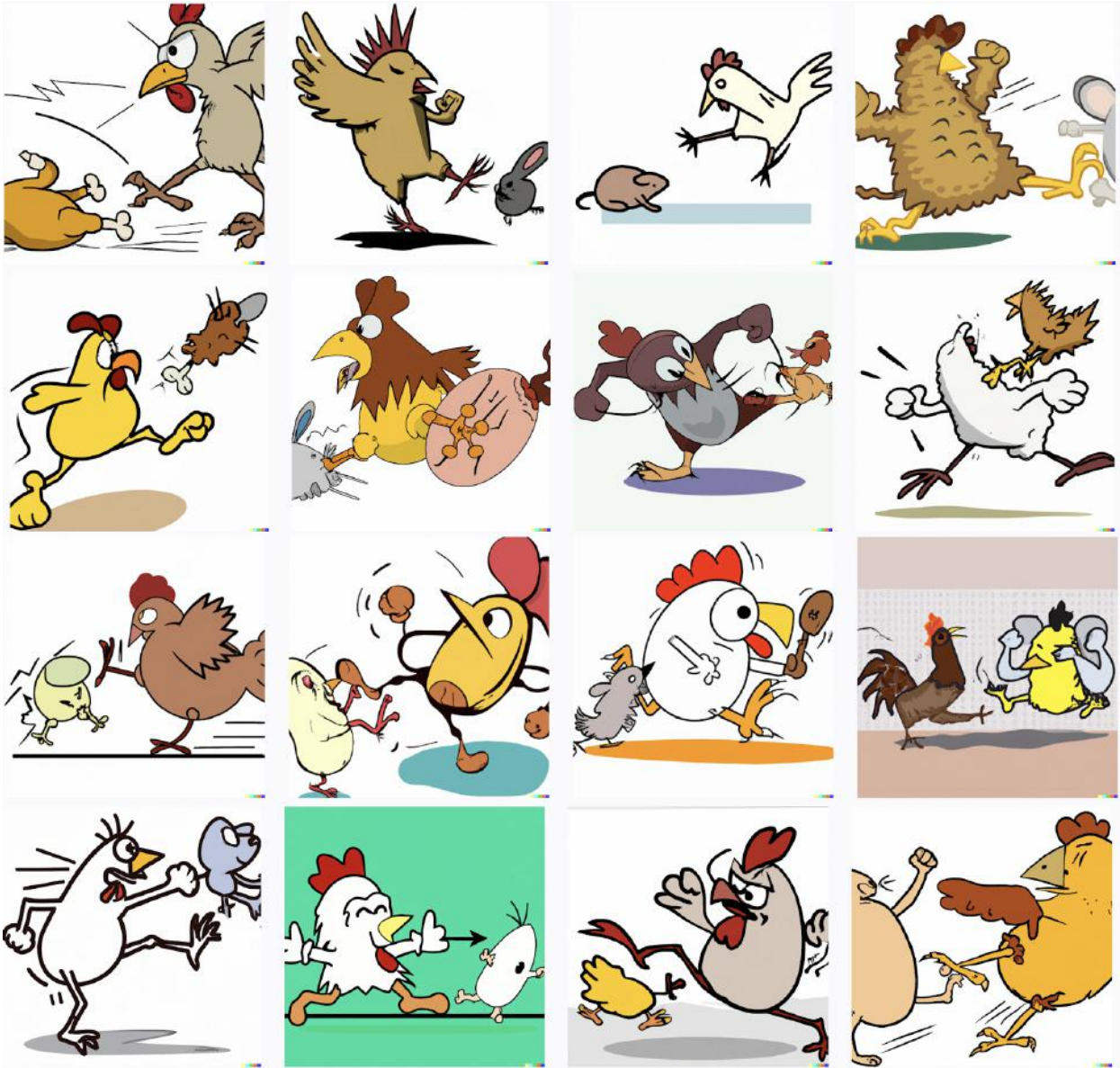

*The sheep is kicking the cat*

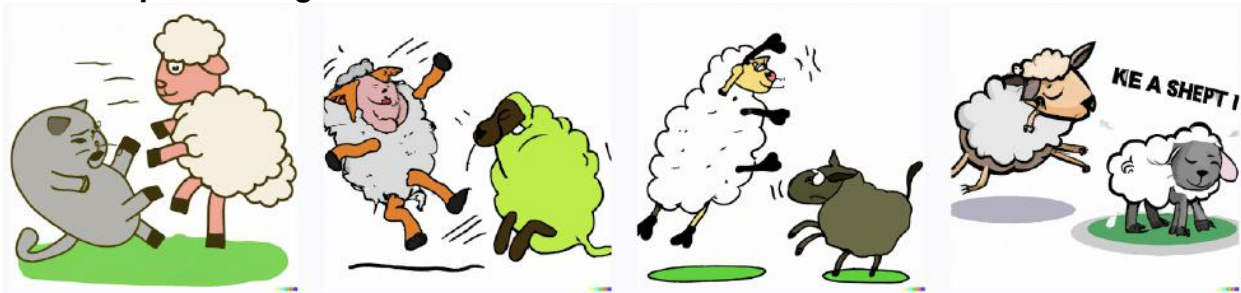

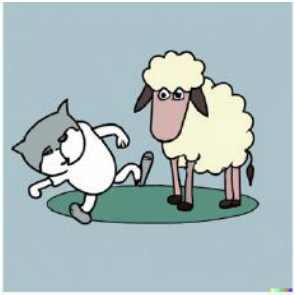 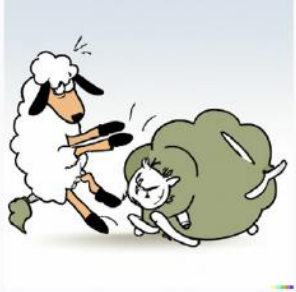 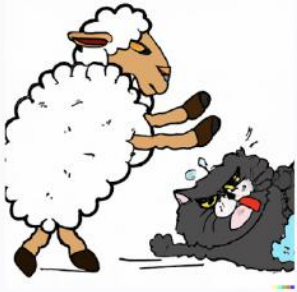 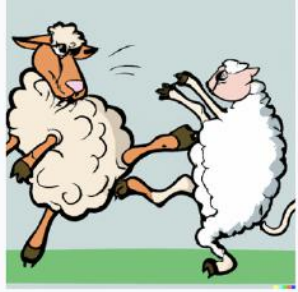
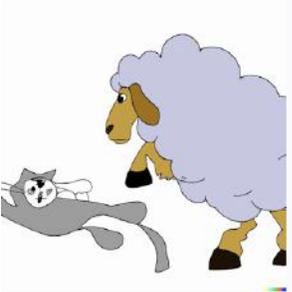 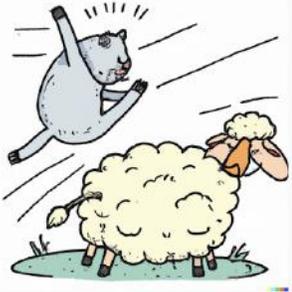 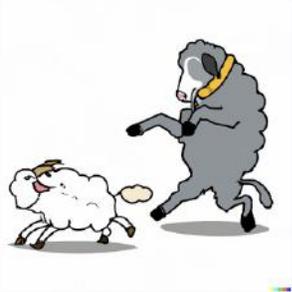 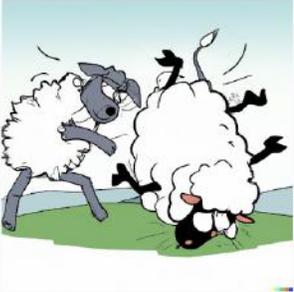
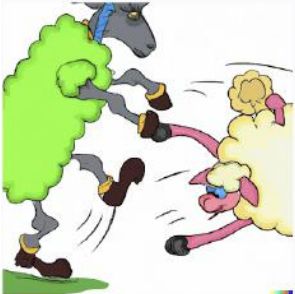 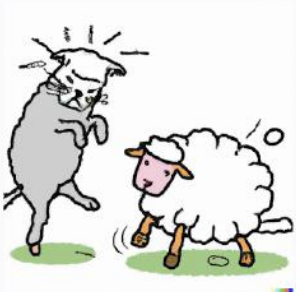 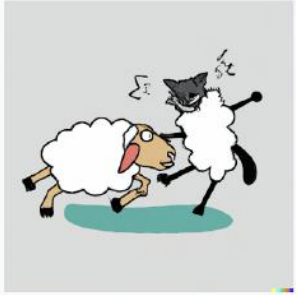 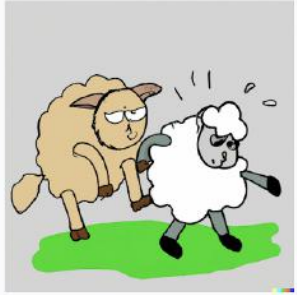
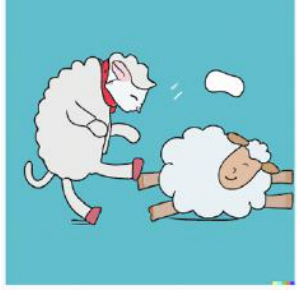 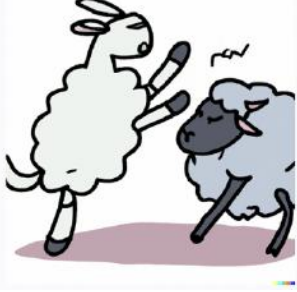 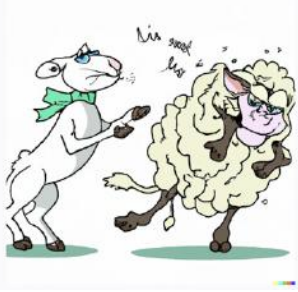 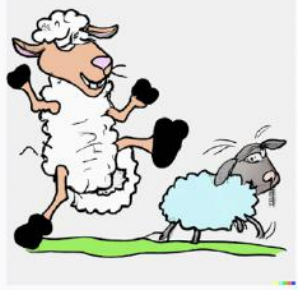